\pdfoutput=1

\documentclass[10pt,journal,compsoc]{IEEEtran}
\usepackage{times}
\usepackage{epsfig}
\usepackage{graphicx}
\usepackage{amsmath}
\usepackage{amssymb}
\usepackage{subfigure}
\usepackage{color}
\usepackage{epstopdf}
\usepackage{algorithm}
\usepackage{algorithmic}
\usepackage{multirow}
\usepackage{booktabs}
\usepackage[inline]{enumitem}
\usepackage{colortbl}
\usepackage[dvipsnames]{xcolor}
\usepackage{url}            

\ifCLASSOPTIONcompsoc
  \usepackage[nocompress]{cite}
\else
  \usepackage{cite}
\fi
\ifCLASSINFOpdf
\else
\fi

\def\eg{\emph{e.g.,}} 
\def\ie{\emph{i.e.,}}

\def\vs{\emph{vs. }}

\def\etal{\emph{et al.}}

\newcommand{\fig}{Fig.}

\begin{document}
%
\title{Leveraging Bottom-Up and Top-Down Attention for Few-Shot Object Detection}
%
%
%
%

\author{Xianyu~Chen, 
        Ming~Jiang, 
        and~Qi~Zhao
\IEEEcompsocitemizethanks{\IEEEcompsocthanksitem X. Chen, M. Jiang, and Q. Zhao are with the Department
of Computer Science and Engineering, University of Minnesota, Minneapolis,
MN, 55455.\protect\\
E-mail: see http://www-users.cs.umn.edu/\~{}qzhao/}%
}
\IEEEtitleabstractindextext{%
\begin{abstract}
Few-shot object detection aims at detecting objects with few annotated examples, which remains a challenging research problem yet to be explored. Recent studies have shown the effectiveness of self-learned top-down attention mechanisms in object detection and other vision tasks. The top-down attention, however, is less effective at improving the performance of few-shot detectors. Due to the insufficient training data, object detectors cannot effectively generate attention maps for few-shot examples. To improve the performance and interpretability of few-shot object detectors, we propose an attentive few-shot object detection network (AttFDNet) that takes the advantages of both top-down and bottom-up attention. Being task-agnostic, the bottom-up attention serves as a prior that helps detect and localize naturally salient objects. We further address specific challenges in few-shot object detection by introducing two novel loss terms and a hybrid few-shot learning strategy. Experimental results and visualization demonstrate the complementary nature of the two types of attention and their roles in few-shot object detection. Codes are available at \url{https://github.com/chenxy99/AttFDNet}.
\end{abstract}

}

\maketitle

\IEEEdisplaynontitleabstractindextext

%
\IEEEpeerreviewmaketitle

\IEEEraisesectionheading{\section{Introduction}\label{sec:introduction}}

%
%
%
%
\IEEEPARstart{H}{umans} can learn novel knowledge from just one or two examples, which is a unique capability that modern artificial intelligence systems yet to develop. Recently, with their remarkable performance driven by large-scale datasets, deep neural networks (DNNs) have dominated the computer vision community. In many applications though, it is labor-intensive and sometimes impractical to collect a large amount of training data and annotations. In object detection, for example, annotating all the possible bounding boxes can be exhausting. Besides, domain-specific applications are generally costly in data collection due to the requirement of expertise.

Few-shot object detection is a trending research topic aiming at training object detectors that generalize well with a small amount of object annotations. Studies have shown that directly applying DNNs designed for big datasets to few-shot object detection tasks often leads to overfitting~\cite{hao:2018:lstd,ze:2020:ctfsod, xin:2020:fsod,bingyi:2019:reweight,xiaopeng:2019:metarcnn,yuxiong:2019:metadet,qi:2019:fsod,yiting:2020:lscn,juanmanuel:2020:once,ting:2019:coae,shafin:2020:asd,siddhesh:2020:waod}. Various learning strategies, such as meta-learning~\cite{bingyi:2019:reweight,xiaopeng:2019:metarcnn,yuxiong:2019:metadet,qi:2019:fsod} and transfer learning~\cite{hao:2018:lstd,ze:2020:ctfsod,xin:2020:fsod}, have been explored to address this issue. Specifically, by combining some state-of-the-art methods gaining significant improvement in object detection, such studies~\cite{yiting:2020:lscn,juanmanuel:2020:once,ting:2019:coae} succeed to boosting the performance of few-shot object detection. More recently, any-shot object detector~\cite{shafin:2020:asd} proposes a realistic setting for the unseen and few-shot novel categories and weakly any-shot object detector~\cite{siddhesh:2020:waod} extend the former any-shot object detector with weakly supervision. However, under extremely few-shot situations, the very limited supervision is still insufficient for learning representative features of objects. Recent few-shot object detection studies~\cite{qi:2019:fsod,bingyi:2019:reweight,xiaopeng:2019:metarcnn} started to use the self-learned attention mechanism due to its effectiveness in many computer vision tasks. Attention often serves as a guidance to the most task-relevant spatial regions where models should assign a high priority when updating their parameters. However, because the learning of attention is dependent on top-down supervision, it could be difficult to train a generalizable attention model when only a small number of training samples are accessible.
Therefore, with fewer training samples, the self-learned top-down attention tends to be less effective.

\begin{figure*}[t]
\centering
\includegraphics[width=1.\linewidth]{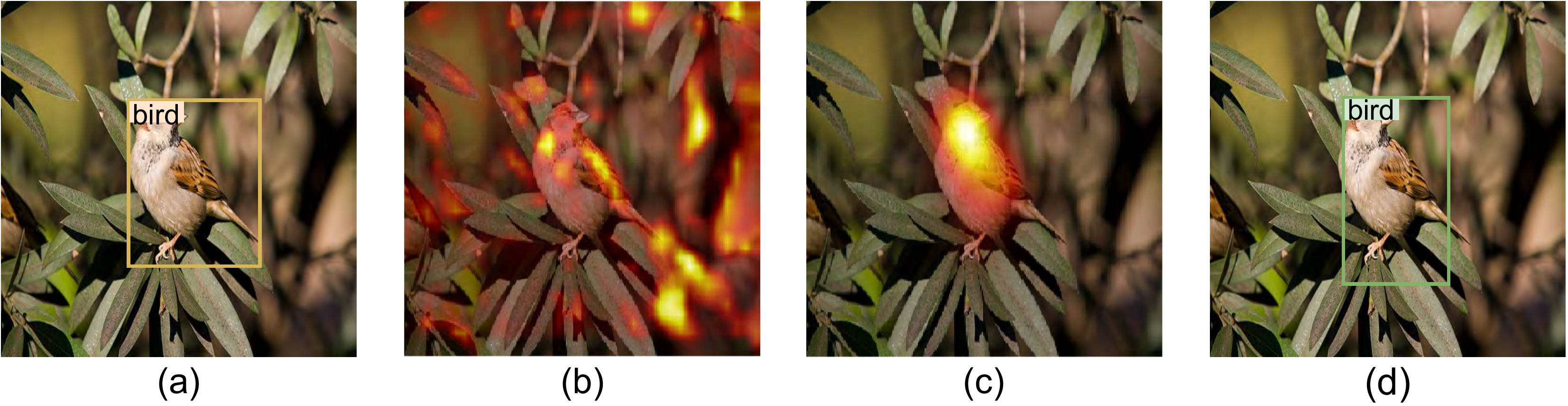}
\caption{In few-shot object detection, due to insufficient supervision, top-down attention learned from object annotations may fail to focus on objects of interest. (a) Input image with the ground-truth bounding box. (b) The top-down attention map. (c) The bottom-up attention map. (d) Detection result of the proposed method. It demonstrates the complementary characteristic of the top-down attention and bottom-up attention, where saliency can provide extra information to compensate the miss of information from top-down attention. We would discuss this characteristic in our qualitative analysis.}
\label{fig:introduce}
\end{figure*}

To address this challenge, we introduce an attentive few-shot object detection network (AttFDNet)
by complementing the top-down attention with bottom-up attention learned from eye-tracking data. The bottom-up attention~\cite{tam:2018:attentivesystem,milica:2012:saliencybias,lori:2012:onsaliency}, also known as saliency, simulates where humans look without influences from tasks. It is directed towards interesting objects that naturally attract attention and can provide supplemental information for the detection of few-shot objects. It has been applied in different tasks, such as mobile robot vision navigation~\cite{chinkai:2010:visionnavigation,chinkai:2011:visionnavigation}, saliency detection~\cite{dapeng:2015:saliencydetection}, salient object detection~\cite{tam:2019:salientod,tam:2017:salientod} and the prediction of where people look~\cite{mengmi:2018:anticipatelook}. As shown in \fig~\ref{fig:introduce}, by combining the top-down attention with bottom-up attention, the proposed object detector can successfully detect and localize few-shot object categories.
Although top-down attention map is spread out, the saliency succeeds to capturing the object of interest and hence helps the detector effectively use the important feature to boost the final detection result.
To support the learning of the proposed attention mechanism in few-shot object detection, we further propose two  concentration losses and a hybrid few-shot learning strategy. The concentration losses allow the detector to boost its ability to discriminate different categories, while the hybrid learning strategy can provide a good initialization to prevent the model from overfitting. Quantitative results on the PASCAL VOC~\cite{mark:2015:voc,mark:2010:voc} dataset show that incorporating bottom-up attention is able to significantly improve the few-shot object detection performance. With qualitative analysis, we further demonstrate the complementary roles of the bottom-up and top-down attention in few-shot object detection.

To sum up, we propose a novel AttFDNet for few-shot object detection. Our contributions are four-fold: 1) we first leverage visual saliency maps as a bottom-up attention mechanism to complement the top-down attention that sometimes fail to capture objects of interest.
2) We propose the object concentration loss and background concentration loss to improve the intra-class agreements and avoid some hard negative anchors introduced to the calculation of the loss.
3) We also design a hybrid few-shot learning strategy that exploits the advantages of both transfer learning and meta-learning.
4) We provide a comprehensive qualitative analysis for the complementary characteristic of top-down attention and bottom-up attention.

We begin with a brief review of the related work in Section~\ref{sec:related_work}. Then, we introduce our AttFDNet in Section~\ref{sec:attentive-object-detection}, which consists of the detailed designs of the object detector, the bottom-up and top-down attention mechanisms, and the objective functions. Next, in Section~\ref{sec:hybrid}, we propose to modify the general one-stage object detector into the hybrid structure suitable for few-shot object detection to prevent from overfitting. In Section~\ref{sec:experiments_and_results}, we perform extensive experiments and show the corresponding quantitative and qualitative results followed by Section~\ref{sec:concluision} concluding the paper.

\section{Related work}
\label{sec:related_work}

With the fast development of deep learning, object detection have achieved significant success with a large amount annotation data. However, it is labor-intensive and sometimes difficult to collect such data and the corresponding annotations. The trending topic, few-shot learning has been developed and obtained remarkable performance in class recognition task. Furthermore, attention plays an important role in many different vision tasks and human eye-tracking data is beneficial for identifying the novel objects.  We discuss these three main topics in this section.

\textbf{Object detection} is a classic computer vision task. Early methods formulate the object detection problem as the classification of a number of candidates sampled from sliding windows~\cite{navneet:2005:hog} or region proposals~\cite{jasper:2013:selectivesearch,lawrence:2014:edgeboxes}. Recently, many DNN-based object detectors have been proposed. Most of these detectors can be generally  categorized as one-stage detectors or two-stage detectors. One-stage detectors, such as CenterNet~\cite{xingyi:2019:pointobject}, YOLO~\cite{joseph:2016:yolo,joseph:2016:yolo2,joseph:2018:yolo3}, SSD~\cite{wei:2016:ssd} and their variants~\cite{cheng-yang:2016:dssd,songtao:2018:rfb}, simultaneously predict the bounding boxes and categories of objects. They are usually more efficient but less accurate. Differently, for the sake of high precision, two-stage detectors explicitly generate class-agnostic region proposals and further classify them into different object categories. R-CNN~\cite{ross:2014:rcnn} and its corresponding variants~\cite{ross:2015:fastrcnn,kaiming:2017:maskrcnn,shaoqing:2017:fasterrcnn} would fall into this category.
Both approaches require intensive supervision to achieve favorable performance, so they are difficult to extend for novel objects with few examples.

\textbf{Few-shot learning} methods~\cite{yaqing:2019:survey-fewshot} have been widely applied in object recognition~\cite{chelsea:2017:maml,weiyu:2019:fewshocls,li:2003:baye-oneshot,koch:2015:siamese,taesup:2018:bmaml,kwonjoon:2019:metaco,jake:2017:prototypical,qianru:2019:metatransfer,flood:2018:relation-network,oriol:2016:matchingnet}, with a focus of addressing the core issue of unreliable empirical risk minimizer.
For object detection, a few different few-shot learning approaches have been proposed. The low-shot transfer detector (LSTD)~\cite{hao:2018:lstd} uses a regularized transfer learning framework to leverage object knowledge from source to target domains.  Context-transformer~\cite{ze:2020:ctfsod} proposes to leverage the object knowledge from source-domain as a guidance and exploit contexts from the training images in the target-domain and hence distinguishes object confusion caused by annotation scarcity.
Few-shot object detection~\cite{xin:2020:fsod} proposes to fine-tune the last layer of the two-stage object detector in the novel training stage with cosine similarity for the box classifier which achieves significant improvement.
RepMet~\cite{leonid:2019:repmet} replaces the standard linear classifier with a distance-based classifier  to allow new few-shot categories to be learned on the fly. Recent research based on meta-learning~\cite{chelsea:2017:maml} has also obtained remarkable performance~\cite{qi:2019:fsod,bingyi:2019:reweight,xiaopeng:2019:metarcnn,yuxiong:2019:metadet,juanmanuel:2020:once}, resulting in better generalization and faster deployment than transfer learning.
Low-shot classification correction network (LSCN)~\cite{yiting:2020:lscn} proposes classification refinement with four different parts (unified recognition, global receptive field, inter-class separation, and confidence calibration) to boost the performance of the overall classes.
Apart from the use of meta-learning, openended centre net (ONCE)~\cite{juanmanuel:2020:once} built on the CenterNet~\cite{xingyi:2019:pointobject} first proposes a new study of incremental few-shot object detection setting, where the new classes are registered incrementally without using the samples from base classes.
Co-attention and co-excitation (CoAE)~\cite{ting:2019:coae}, uses the nonlocal operation~\cite{xiaolong:2018:nolocal} to explore the co-attention embodied in the query-target pair and the squeeze-and-co-excitation scheme~\cite{jie:2018:senet} to emphasize the correlated feature channels to uncover the relevant proposals. Then it proposes a proposal ranking that the most relevant proposals to the query would appear in the top portion of the ranking list by a margin-based ranking algorithm.
Any-shot object detection~\cite{shafin:2020:asd} proposes a realistic setting for the unseen and few-shot novel categories and trains them simultaneously in the same framework. The few-shot object detection is the special case of this framework. Furthermore, weakly-supervised any-shot object detection~\cite{siddhesh:2020:waod} introduces the weakly-supervision into the any-shot object detector. DID~\cite{xianyu:2020:did} proposes to address the continuous low-shot object detection problem, which is different from our
aim and does not introduce any attention mechanism to the object detectors.
Most of these methods are based on meta-learning that is known to be very sensitive to novel examples and may cause a performance drop on base categories. They also fail to work in extremely few-shot scenarios. To improve the detection performance for both base and novel categories, the proposed attentive few-shot object detector adopts a hybrid learning strategy with both transfer learning and imprinting.

\textbf{Attention} has been widely used in the design of neural networks for many vision tasks~\cite{yue:2019:gcnet,jan:2015:speechrecog,zhenyang:2018:videolstm,kelvin:2015:image-caption}. The attention mechanism grants models the ability to focus on more important spatial locations or feature channels. It is noteworthy that even without the explicit supervision from human eye-tracking data, many attention models can learn where to focus simply by optimizing the task objectives. Supervised with a sufficient amount of annotations, the self-learned top-down attention can perform reasonably well in many vision tasks.
Recent literature~\cite{yue:2019:gcnet,xiaolong:2018:nolocal} have exhaustively experimented the use of attention in object detection related tasks. The recently proposed few-shot object detectors~\cite{qi:2019:fsod,bingyi:2019:reweight,xiaopeng:2019:metarcnn} have also demonstrated the effectiveness of top-down attention. Specifically, YOLO-Low-Shot~\cite{bingyi:2019:reweight} and Meta~R-CNN~\cite{xiaopeng:2019:metarcnn} both use a meta-model to produce reweighting attentive vectors for each specific category from the support set. They apply the attentive vectors on intermediate query features to obtain better detection results. Further, Fan~\etal~\cite{qi:2019:fsod} propose an Attention-RPN that calculates the depth-wise cross correlation between support features and query features to improve the few-shot detection performance. While these studies learn attention in a top-down manner, the computed attention vectors or maps often fail to highlight the correct objects of interest, resulting in suboptimal model performance and lack of interpretability. In this work, we demonstrate that bottom-up attention could play an important role in the few-shot object detection task to address such issues, which has been shown effective in other vision tasks (\eg~image captioning~\cite{kelvin:2015:image-caption}, machine translation~\cite{jan:2015:speechrecog}, and action recognition~\cite{zhenyang:2018:videolstm}).

\begin{figure}[t]
\centering
\includegraphics[width=1.\linewidth]{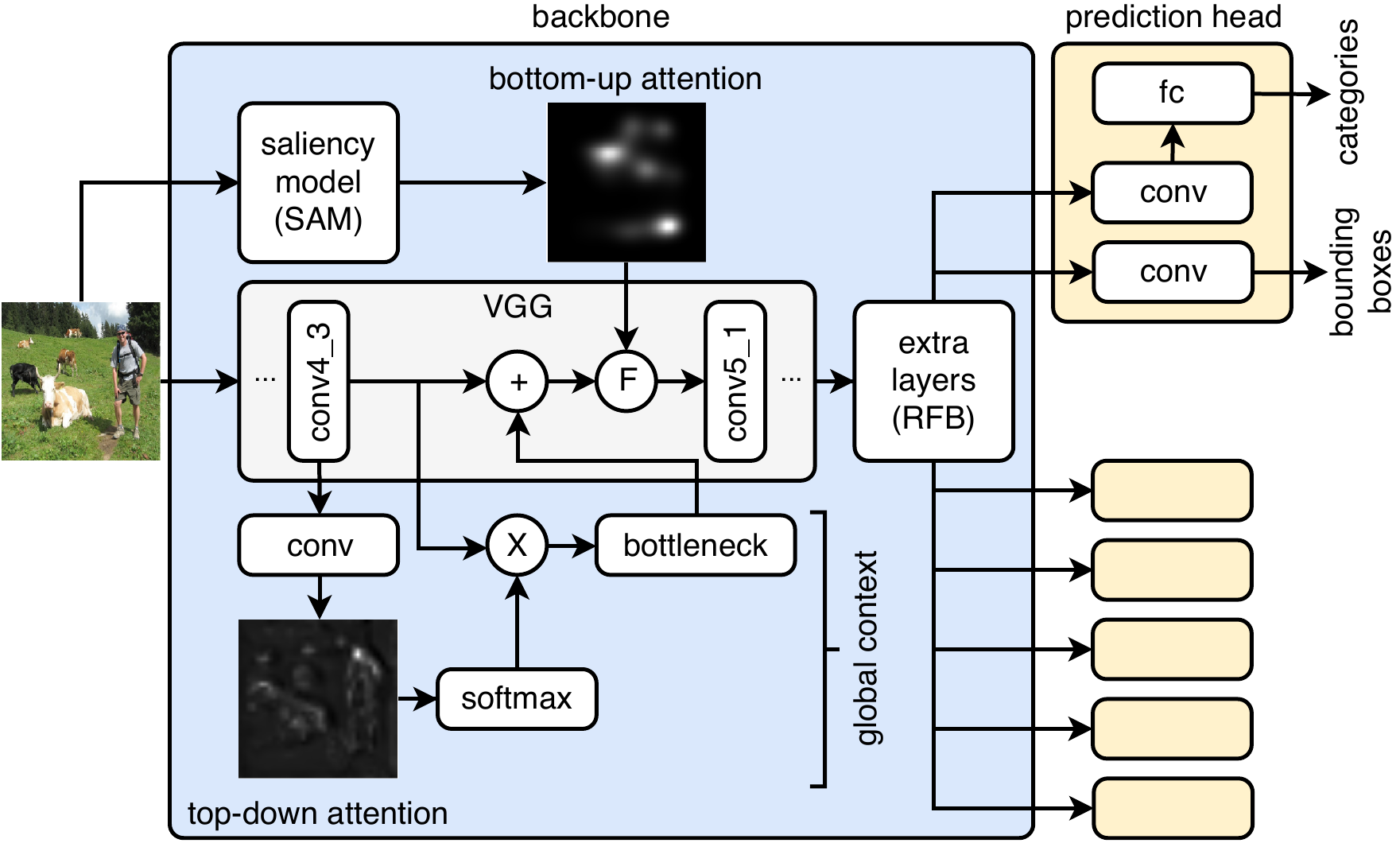}
\caption{The network architecture of the proposed attentive few-shot object detector. First, we use the saliency model to generate the bottom-up attention for a given image. Then we send the image to the backbone, and use the generated bottom-up attention as well as the top-down attention through the backbone to provide a guidance of the specific spatial feature map. Last, we arrive to the six prediction heads to get the corresponding detection results related to the localization and category of an object. The backbone of the network is highlighted in blue, while the six prediction heads are highlighted in yellow.}
\label{fig:framework}
\end{figure}

\section{Attentive Few-Shot Object Detection Network}
\label{sec:attentive-object-detection}

Training a few-shot object detector consists of two stages. In the \textbf{base-training} stage, a general object detector is trained on a large set of annotated images to detect a number of base categories. Next, in the \textbf{novel-training} stage, several novel object categories are added to the training data, each with only $K$ annotations. The goal of few-shot object detection is to train a $K$-shot object detector by making use of the already trained base detector and the new data to detect all objects of the base and novel categories. In this section, we first focus on the general base-training stage and present the design of an attentive few-shot object detector. We will then introduce a novel concentration loss to address particular challenges in the novel-training stage.

The attentive few-shot object detector is composed of a one-stage object detection backbone and a number of prediction heads. In particular, the backbone network leverages both bottom-up and top-down attention for object detection. While the use of a self-learned top-down attention has been proven effective in conventional object detection tasks, in few-shot object detection, the performance of top-down attention is still limited with the insufficiency of training data and the misalignment with human visual attention. Thus, our method differentiates itself from the previous works by highlighting the importance of bottom-up attention in few-shot object detection. Although the network architecture is applicable to object detection in general, it is particularly useful in the few-shot context because of the complementary bottom-up attention mechanism that naturally detects regions of interest without requiring top-down supervision.

\subsection{One-Stage Object Detection}\label{subsec:onestage-detector}

The proposed object detector flexibly integrates various kernels and dilated convolution layers into an SSD-style detector~\cite{songtao:2018:rfb} (see \fig~\ref{fig:framework}). It is composed of a visual encoder (\ie~VGG Net~\cite{liang:2018:deeplabvgg,kares:2015:vgg}), two attention pathways (\ie~bottom-up attention and top-down attention), and six prediction heads that detect objects at different scales. Each detector head uses a convolutional layer to predict the bounding boxes, and another convolutional layer followed by a fully-connected layer to predict the object categories.

\subsection{Bottom-Up and Top-Down Attention}\label{subsec:modelatt}

An intrinsic challenge of few-shot object detection is that object detectors hardly find correct regions where the objects can be detected. The proposed attention pathways address this challenge by assigning different weights to the spatial locations, so that important features do not get ignored. As illustrated in \fig~\ref{fig:introduce}, bottom-up attention and top-down attention detect important image regions in different ways. On the one hand, the detector can learn model attention by itself from the bounding box annotations, which has been commonly accepted by many studies due to its efficiency and effectiveness. On the other hand, bottom-up attention computed by a saliency prediction algorithm is not only interpretable but also applicable in a similar way to prioritize features. In the following, we introduce the design of top-down attention and then present how bottom-up attention can be incorporated to further improve the detector's performance.

On the one hand, the design of our top-down attention model is inspired by the global context (GC) block~\cite{yue:2019:gcnet}, which benefits from the simplified nonlocal block~\cite{xiaolong:2018:nolocal} and the squeeze-excitation (SE) block~\cite{jie:2018:senet}. It is a lightweight network capturing long-range dependencies for the global representation of a scene effectively.
To compute the top-down attention, we extract the $C$-dimensional feature maps $\mathbf{y}\in \mathbb{R}^{C\times H \times W}$ from the \texttt{conv4\_3} layer, where $\mathbf{y}_{:,i,j}\in \mathbb{R}^{C}$ represents the local feature vector at the ${i,j}$-th pixel of the feature maps. The feature maps $\mathbf{y}$ is fed to a convolution layer $\mathbf{W}_k$ to compute a soft attention map
\begin{equation}\label{equ:global_att_hm}
  \mathbf{h}_{i,j} = \frac{e^{\mathbf{W}_k\mathbf{y}_{:,i,j}}}{\sum_{n=1}^{H}\sum_{m=1}^{W}e^{\mathbf{W}_k\mathbf{y}_{:,n,m}}}.
\end{equation}
The attention map $\mathbf{h}$ is used to model the spatial response from the top-down attention. It is multiplied with the feature maps $\mathbf{y}$ to compute the weighted global features to capture the long-range dependencies for the global representation of a scene:
\begin{equation}\label{equ:hm_tensor_y}
  \mathbf{y}' = \mathbf{y}\star \mathbf{h},
\end{equation}
where $\star$ represents the tensor multiplication. More specifically, we formulate the expression of $\mathbf{y}'$: $\mathbf{y}' = \sum_{i,j}\mathbf{y}_{:,i,j} \mathbf{h}_{i,j}$.
Hence $\mathbf{y}'$ models the global context of this specific feature map.

The global feature transform is specifically designed for the top-down attention to capture the channel-wise dependencies. As a lightweight attention block, it is easier to fine-tune with few training samples. Then, we put the weighted global features $\mathbf{y}'$ through a bottleneck transform that consists of a convolution layer $\mathbf{W}_{v1}$, a normalization layer (LN), a ReLU operation, and a convolution layer $\mathbf{W}_{v2}$, sequentially. An element-wise addition is used to fuse the bottleneck output with the original features in a residual form:
\begin{equation}\label{equ:fusion}
  \mathbf{z} = \mathbf{y} + \mathbf{W}_{v2}\text{ReLU}\big(\text{LN} (\mathbf{W}_{v1}\mathbf{y}')\big).
\end{equation}

To further improve the object detection performance, we compute the saliency map of the input image $\mathbf{x}$, using a bottom-up attention model
denoted as $g(\mathbf{x};\varphi)$, where $\varphi$ represents the parameters of this model.
The computed saliency map is transformed and multiplied with the fused features $\mathbf{z}$ as
\begin{equation}\label{equ:merge}
  \mathbf{z}' = \mathbf{z} \odot \ln(\epsilon+g(\mathbf{x};\varphi)),
\end{equation}
where $\epsilon$ is a hyper-parameter, which is used to control the influence of the bottom-up attention and $\odot$ represents the channel-wise multiplication.

Finally, the attended features $\mathbf{z}'$ are used as the input to the next convolution layer of the object detection network and the remaining backbone is unchanged.

\subsection{Objectives for Base and Novel Detectors}\label{sec:loss_func_base}

\textbf{Base-training loss function.} The loss commonly used in object detection~\cite{ross:2015:fastrcnn,wei:2016:ssd} contains a smooth-$L_1$ term for bounding box regression (\ie~ $L_{\text{bbox}}$) and a multi-class cross entropy $L_{\text{cls}}$ for object classification~\cite{songtao:2018:rfb,wei:2016:ssd}. It can be denoted as
\begin{equation}\label{equ:base-detection-loss}
  L(\theta) = \frac{1}{N}\Big(L_{\text{cls}}(\theta)+\alpha L_{\text{bbox}}(\theta)\Big),
\end{equation}
where $N$ is a default number of matched bounding boxes, and $\alpha$ is the hyper-parameter that balances the weights of the two loss terms. In the base-training stage, the proposed object detector is trained using this general loss function.


\textbf{Novel-training loss function.}
In the novel-training stage, a number of challenges arise due to the limited number of training examples. To address these challenges, we design two concentration losses for few-shot object detection: the object-concentration loss and the background-concentration loss. They allow the object detector to better classify positive/negative anchors (\ie~bounding box candidates) in few-shot object detection.
In brief, the object concentration loss helps the features corresponding to the same object get together.

\textit{1. Object-concentration loss.} In few-shot object detection, the commonly used cross-entropy loss sometimes fails to distinguish similar categories (\eg~cats~\vs~dogs). The reason is that the number of samples is not enough for the fully-connected layer to learn more discriminative parameters and difficult to train the VGG~\cite{liang:2018:deeplabvgg,kares:2015:vgg} and extra layers~\cite{songtao:2018:rfb} to encode more representative features for different categories. To better classify different few-shot object categories, we propose an object-concentration loss that \textbf{maximizes} the cosine similarity between positive anchors (\ie~those having a sufficiently large IoU with the ground-truth bounding box) of the same category. It pushes the features and their corresponding weights in the fully-connected layer closer to improve their intra-class agreements.

The object-concentration loss measures the cosine similarity between the final convolutional layer features and the weights of the fully-connected layer. Specifically, we denote the features of the prior anchors as $\mathbf{f}_i,~1\leq i\leq N_{\text{anchor}}$, where $N_{\text{anchor}}$ is the number of prior anchors in RFB~\cite{songtao:2018:rfb}. We also denote the weights of the final fully-connected layer as $\mathbf{w}_j,~j=1,2,\cdots,N_{\text{cls}}$, where $N_{\text{cls}}$ is the number of categories. Both the features and the weights are normalized into unit vectors~\cite{hang:2017:imprinting}.
Next, we define an indicator function $I_{i,j},~1\leq i\leq N_{\text{anchor}},~j=1,2,\cdots,N_{\text{cls}}$ to represent whether the $i^{th}$ prior anchor belongs to the $j^{th}$ category. The object-concentration loss for the positive anchors is denoted as
\begin{equation}\label{equ:object-concentration-loss}
  L^{+}_{\text{conc}}(\theta)=-\frac{\sum_{i=1}^{N_{\text{anchor}}}\sum_{j=1}^{N_{\text{cls}}}I_{i,j}\mathbf{w}_j^T\mathbf{f}_i}{\sum_{i=1}^{N_{\text{anchor}}}\sum_{j=1}^{N_{\text{cls}}}I_{i,j}}.
\end{equation}

\begin{figure*}[t]
\centering
\includegraphics[width=1.0\linewidth]{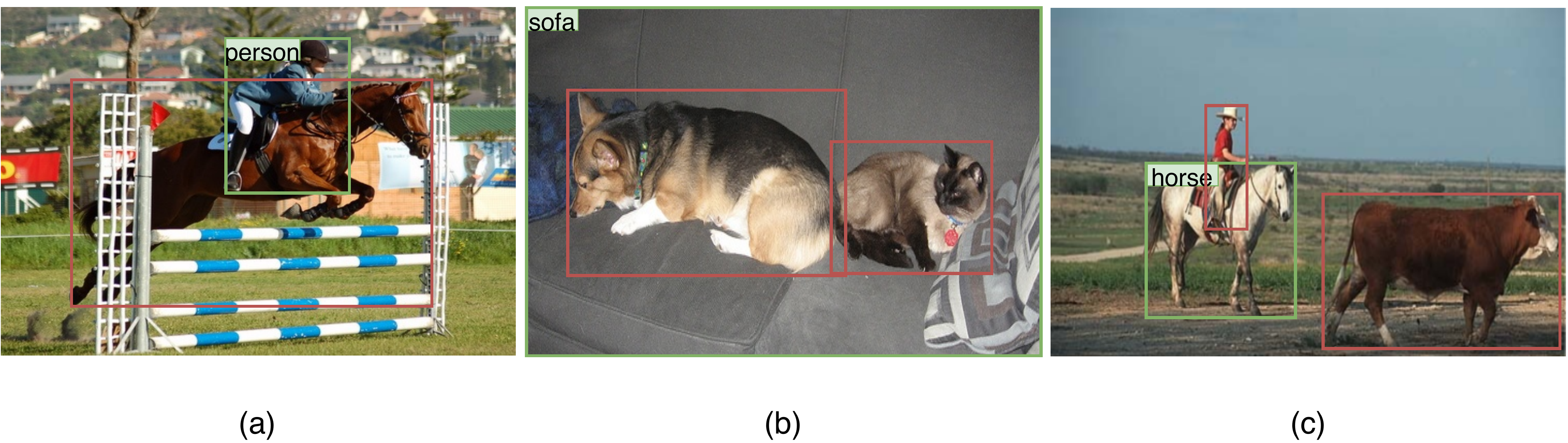}
\caption{A unique challenge of few-shot object detection is that not all bounding boxes in the novel images are annotated for training. The green bounding boxes indicate few-shot annotations and the red bounding boxes represent unannotated objects. With conventional training methods, such incomplete annotations would cause performance degradation.}
\label{fig:concentration-loss}
\end{figure*}

\textit{2. Background-concentration loss.} A unique challenge of few-shot object detection is the unavailability of complete annotations \ie~both the base and novel objects can remain unlabelled in the training images. On the one hand, novel objects (\eg~\textit{horse} in \fig~\ref{fig:concentration-loss}a) can be unlabeled in the base training set. On the other hand, base objects (\eg~\textit{person}, \textit{dog}, \textit{cat} in \fig~\ref{fig:concentration-loss}b-c) or objects of different novel categories (\eg~\textit{cow} in \fig~\ref{fig:concentration-loss}c) can also be unlabelled in the novel training set. In applications, there are two practical reasons of such incomplete annotations: (1) The two datasets are developed separately with different focuses. (2) Completely annotate the novel dataset may significantly increase the cost of data collection.
Because of the incomplete annotations, with hard negative example mining~\cite{wei:2016:ssd}, the anchors corresponding to unlabelled objects are likely to be used as negative examples. Training a detector with such examples can lead to a catastrophic detection performance.

To tackle this problem, we define a background-concentration loss to \textbf{minimize} the cosine similarity between the feature corresponding to the selected hard negative anchors and the weights for background in fully-connected layer. We represent the fully-connected layer weights corresponding to the background category as $\mathbf{w}_0$, and define a indicator function $I_{i},~1\leq i\leq N_{\text{anchor}}$ to represent whether the $i^{th}$ prior anchor is indicated as a background during hard negative example mining. The background-concentration loss is defined as

\begin{equation}\label{equ:background-concentration-loss}
  L^{-}_{\text{conc}}(\theta)=\frac{\sum_{i=1}^{N_{\text{anchor}}}I_{i}\mathbf{w}_0^T\mathbf{f}_i}{\sum_{i=1}^{N_{\text{anchor}}}I_{i}}.
\end{equation}




Taking the base-training loss and both concentration losses into account, the final loss function for novel-training is defined as
\begin{equation}
\begin{aligned}
\label{equ:novel-detection-loss}
  L(\theta) &=  \frac{1}{N}\Big(L_{\text{cls}}(\theta)
  + \alpha L_{\text{bbox}}(\theta)\Big)\\
  &+ \beta L_{\text{conc}}^{+}(\theta) + \eta L_{\text{conc}}^{-}(\theta)
   + \gamma L_{\text{dist}}(\theta),
\end{aligned}
\end{equation}
where $L_{\text{dist}}(\theta)$ is the knowledge distillation loss that keeps the balance between the base categories and the novel categories~\cite{geoffrey:2015:distill,konstantin:2017:incremental_learning}. Specifically, following the incremental learning strategy~\cite{konstantin:2017:incremental_learning}, we set the knowledge distillation loss $L_{\text{dist}}(\theta)$ as the combination of the logits and the regression outputs between the base categories results from the base detector and the corresponding base categories results from the few-shot novel detector. Following~\cite{konstantin:2017:incremental_learning}, we use $L2$ loss for regression outputs as it demonstrates more stable training and performs better.
The hyper-parameters $\alpha$, $\beta$, $\eta$, $\gamma$ determine the weights of the corresponding loss terms.

\section{Hybrid Few-Shot Learning Strategy}
\label{sec:hybrid}

Conventional few-shot learning methods are based on either transfer learning or meta-learning. In this work, we propose a hybrid few-shot learning strategy that exploits the advantages of both while resolving their problems. In particular, with the proposed network architecture, we first train a base object detector on a large dataset that provides sufficient annotations of base categories. In the novel-training stage, to make the best use of the knowledge learned by the base detector, we initialize the parameters of the novel object detector using parameters from the base object detector and a imprinting initialization method. The parameter initialization also allows the novel object detector to overcome overfitting incurred from the lack of training data.

\begin{figure}[t]
\centering
\includegraphics[width=0.45\textwidth]{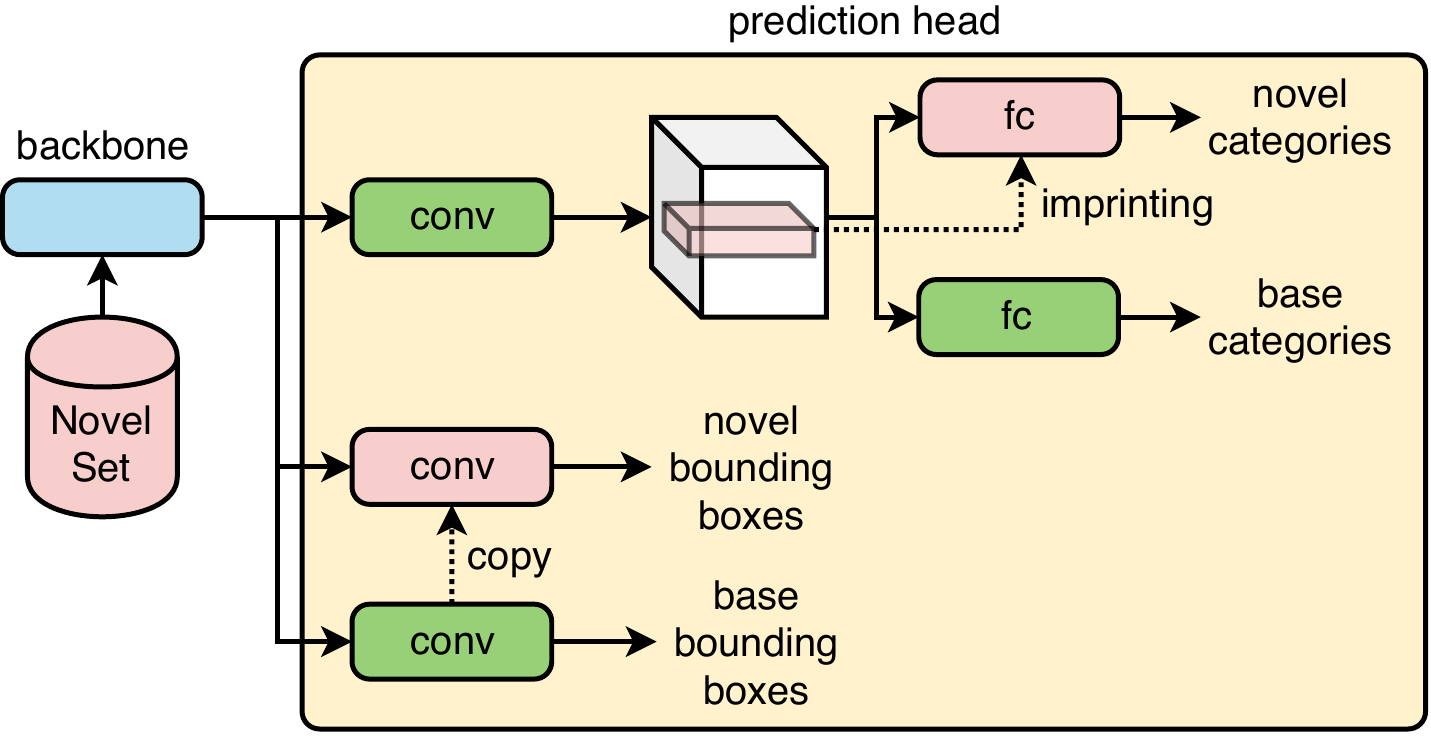}
\caption{Parameter initialization for the novel object detector. In this figure, we show the comprehensive procedure to initialize the novel object detector from base object detector. We use the green blocks to represent the parameters from the prediction heads in the base object detector while the red blocks are the parameters for the prediction heads of the novel object detector. The parameters of novel bounding boxes can be directly copied from the already learned corresponding parameters from the base object detector. The parameters of the fully-connected layer for the novel categories can be initialized from imprinting method~\cite{hang:2017:imprinting,xianyu:2020:did}.
}
\label{fig:novel-training-initialization}
\end{figure}

As shown in \fig~\ref{fig:novel-training-initialization}, since the novel object detector is extended from the base object detector, the backbone (blue) and the prediction head layers corresponding to the base object categories (green) can be directly initialized with the parameters of the base object detector. However, given the new object categories added, the novel detector has additional parameters (red) for predicting the novel object categories and their bounding box positions. These additional parameters need to be initialized before novel-training.

We adopt two different strategies to initialize the convolutional layers for bounding box regression and the fully-connected layers for object classification. For the \emph{bounding box regression}, since the convolutional layer is object-agnostic and only encodes information for detecting boundaries, we can directly copy the already learned parameters from the base object detector to the novel one. For the \emph{object classification}, we initialize the final fully-connected layer following the imprinting method~\cite{hang:2017:imprinting,xianyu:2020:did}. Because the penultimate convolutional layer of the base object detector outputs discriminative features for different object categories, we can feed the novel images to the base object detector, and extract the features from its penultimate layer. After a normalization, we compute the average of these features for each novel category, and use the averaged feature vector as the weights of the fully-connected layer.

With this hybrid strategy, the novel object detector can be properly initialized, so that the novel learning will not overfit the support set and forget the base categories. It is fine-tuned on the support set containing all the novel data and a small portion of samples from the base categories, until the object detection performance converges for all the base and novel categories.

\section{Experiments and Results}
\label{sec:experiments_and_results}

We conduct experiments and ablation studies to demonstrate the effectiveness of the proposed method. This section reports experimental details and results.

\subsection{Dataset and Settings}

The proposed few-shot object detector is trained and evaluated on the widely-used PASCAL VOC~\cite{mark:2015:voc,mark:2010:voc} dataset. Following the common practice~\cite{ross:2015:fastrcnn,wei:2016:ssd,joseph:2016:yolo2}, we train our detector on the $2007$ and $2012$ training and validation sets, and use the $2007$ test set for evaluation. Similar to the settings of YOLO-Low-Shot~\cite{bingyi:2019:reweight}, Meta~R-CNN~\cite{xiaopeng:2019:metarcnn} and MetaDet~\cite{yuxiong:2019:metadet}, we evaluate the trained detectors with three splits, each selecting $5$ random novel categories and using the remaining $15$ as the base categories. The three different novel class sets are (\emph{bird}, \emph{bus}, \emph{cow}, \emph{mbike}, \emph{sofa}), (\emph{aero}, \emph{bottle}, \emph{cow}, \emph{horse}, \emph{sofa}) and (\emph{boat}, \emph{cat}, \emph{mbike}, \emph{sheep}, \emph{sofa}), respectively. All the annotations for the base categories are used in the base-training stage, while only $K$ bounding boxes for each novel class and $3K$ bounding boxes for each base category are used in the novel-training stage~\cite{xiaopeng:2019:metarcnn}. The $K$-shot object detection is performed based on extremely few-shot cases $K=(1,2,3)$ across all three base/novel splits.

We also evaluate our methods on  MS~COCO~\cite{tsung:2014:mscoco}, the widely-accepted object detection benchmark. With $118$k training images (train2017) and $5$k validation images (val2017), MS~COCO has a more diverse set of $80$ categories compared with PASCAL VOC. In our experiments, we choose the same $20$ categories as PASCAL VOC~\cite{mark:2015:voc,mark:2010:voc}, and use them and remaining $60$ categories as the novel categories and base categories.

We further consider the cross-benchmark setting to transfer knowledge from COCO to PASCAL VOC. The few-shot object detector is trained with the 60 base categories of MS~COCO and evaluated on the 20 novel categories of PASCAL VOC.

\subsection{Implementation Details}

The resolution of the input is $300\times 300$ pixels following the RFB~Net~\cite{songtao:2018:rfb}. We adopt the Saliency Attentive Model (SAM)~\cite{marcella:2018:sam} to predict the bottom-up attention map. SAM was trained on the SALICON dataset~\cite{ming:2015:salicon} with all the parameters are frozen while training the object detector.
We set the hyper-parameter $\alpha=1$ following the RFB~Net~\cite{songtao:2018:rfb,wei:2016:ssd}, and set $\gamma=1$ following the common practice~\cite{konstantin:2017:incremental_learning}.

In the base-training phase, we use the Adam~\cite{diederik:2015:adam} optimizer with $\beta_1=0.9$, $\beta_2=0.999$ and $L_2$ weight-decay $0.0005$. The batch size is $64$. The base object detector is trained for $150$ epochs, using a step-wise learning rate decay: $4\times 10^{-4}$ by epoch $90$, $4\times 10^{-5}$ by epoch $120$, and $4\times 10^{-6}$ by epoch $140$. Following the RFB~Net~\cite{songtao:2018:rfb}, we also use the warmup strategy to stabilize the the training.

In the novel-training stage, we use the SGD optimizer with momentum $0.9$. The novel object detector is fine-tuned for $600$ epochs, using a step-wise learning rate decay: $2\times 10^{-3}$ by epoch $250$, $2\times 10^{-4}$ by epoch $400$, and $2\times 10^{-5}$ by epoch $500$.

\subsection{Evaluation Results}

\noindent\textbf{PASCAL VOC.} First, we present quantitative evaluations of the proposed method.

\underline{\smash{Comparison with baseline.}} To demonstrate the advantages of the introduced approach, we compare two variants of our method with a baseline method. The baseline, namely \textbf{RFB-ft-full}, takes a two-phase training strategy to directly fine-tune the top-down attention-based RFB object detector~\cite{songtao:2018:rfb}. It first uses the base categories to train the detector, and then uses the combination of the base categories and novel categories to fine-tune the detector until it fully converges. The two variants of our method use different saliency models as bottom-up attention. \textbf{AttFDNet (BU'+TD)} uses the BMS~\cite{jianming:2013:bms} saliency model and \textbf{AttFDNet (BU+TD)} uses the SAM~\cite{marcella:2018:sam} saliency model. Both of them utilize the proposed concentration losses and the hybrid learning strategy.
Compared with the baseline (RFB-ft-full), AttFDNet (BU+TD) improves the performance significantly (\eg~+$135\%$ for $1$-shot on Split 2 and +$169\%$ for $2$-shot on Split 3). The improvements may come from multiple components of our proposed methods (\ie~the use of bottom-up attention, concentration losses and the hybrid learning strategy). Between SAM and BMS saliency models, the two detectors' performances are similar. The AttFDNet (BU'+TD) model is also significantly better than the RFB-ft-full baseline. It demonstrates the robustness and generalizability of the proposed method. It also suggests that the performance gain is not from the external training on the SALICON dataset, since the BMS model is not data-driven. 

\begin{table*}[t]
\begin{center}
\caption{Mean average precision (mAP) on three different splits of novel categories. We compare the state-of-the-art methods with a baseline (RFB-ft-full) and two variants of our methods: AttFDNet (BU'+TD) and AttFDNet (BU+TD).
}
\label{table:sota-compare}
\scriptsize
\begin{tabular}{{l}*{12}{c}}
\toprule
 & \multicolumn{3}{c}{Split 1 mAP} & \multicolumn{3}{c}{Split 2 mAP} & \multicolumn{3}{c}{Split 3 mAP} & \multicolumn{3}{c}{Average mAP}\\
\cmidrule(lr){2-4} \cmidrule(lr){5-7} \cmidrule(lr){8-10} \cmidrule(lr){11-13}
Method/Shot & 1 & 2 & 3 & 1 & 2 & 3 & 1 & 2 & 3 & 1 & 2 & 3\\
\midrule
LSTD (YOLO)-full~\cite{hao:2018:lstd} & 8.2 & 11.0 & 12.4 & 11.4 & 3.8 & 5.0 & 12.6 & 8.5 & 15.0 & 10.7& 7.8& 10.8 \\
YOLO-Low-Shot~\cite{bingyi:2019:reweight} & 14.8 & 15.5 & 26.7  & 15.7 & 15.3 & 22.7 & 21.3 & 25.6 & 28.4 & 17.3 & 18.8 & 25.9   \\
Meta R-CNN~\cite{xiaopeng:2019:metarcnn} & 19.9 & 25.5 & 35.0 &  10.4 & 19.4 & 29.6  & 14.3& 18.2 & 27.5 & 14.9 & 21.0 & 30.7\\
MataDet~\cite{yuxiong:2019:metadet} & 18.9 & 20.6 & 30.2 & 21.8 & 23.1 & 27.8 & 20.6 & 23.9 & 29.4 & 20.4& 22.5& 29.1 \\
\midrule
RFB-ft-full  & 15.6 & 22.8 & 25.2 & 6.8 & 8.1 & 12.6 & 7.2 & 10.8 & 17.8 & 9.9 & 13.9 & 18.5 \\
AttFDNet (BU'+TD) & 29.1 & 34.0 & 35.0 & 16.2 & 20.9 & 22.6 & 21.5 & 28.8 & 31.0 & 22.3 & 27.9 & 29.5  \\
AttFDNet (BU+TD) & 29.6 & 34.9 & 35.1 & 16.0 & 20.7 & 22.1 & 22.6 & 29.1 & 32.0 & 22.7 & 28.2 & 29.7  \\
\bottomrule
\end{tabular}
\end{center}
\end{table*}

\underline{\smash{Comparison with the state of the art.}} The results of our proposed AttFDNet (BU+TD) is also compared with four state-of-the-art few-shot object detectors, LSTD~\cite{hao:2018:lstd}, YOLO-Low-Shot~\cite{bingyi:2019:reweight}, Meta~R-CNN~\cite{xiaopeng:2019:metarcnn} and MetaDet~\cite{yuxiong:2019:metadet}. For a fair comparison, we use the same data sampling method as Meta R-CNN~\cite{xiaopeng:2019:metarcnn} does. As shown in Table~\ref{table:sota-compare}, the AttFDNet (BU+TD) outperforms LSTD (YOLO)-full~\cite{hao:2018:lstd} in all the $9$ cases (\eg~+$217\%$ for $2$-shot on Split 1; +$242\%$ for $2$-shot on Split 3), YOLO-Low-Shot~\cite{bingyi:2019:reweight} in $8$ out of the $9$ cases (\eg~+$100\%$ for $1$-shot on Split 1; +$35\%$ for $2$-shot on Split 2), the Meta~R-CNN~\cite{xiaopeng:2019:metarcnn} in $8$ out of the $9$ cases (\eg~+$54\%$ for $1$-shot on Split 2 and +$60\%$ for $2$-shot on Split 3) and the MetaDet~\cite{yuxiong:2019:metadet} in $6$ out of the $9$ cases (\eg~+$69\%$ for $2$-shot on Split 1; +$22\%$ for $2$-shot on Split 3). These improvements demonstrate the effectiveness of the proposed method in few-shot object detection.
Notably, with a relatively low-resolution input (\ie~$300\times 300$ pixels), our method is also more cost-efficient than YOLO-Low-Shot ($416\times 416$ pixels) and Meta~R-CNN ($800\times 600$ pixels).

\begin{table*}[t]
\begin{center}
\caption{Average precision (AP) and mAP on base and novel categories of the second and third base/novel splits. Under $2$-shot scenario on the second and third splits of base/novel categories, we compare the state-of-the-art methods with a baseline (RFB-ft-full) and two variants of our methods: AttFDNet (BU'+TD) and AttFDNet (BU+TD).
}
\label{table:detail-compare}
\resizebox{1\textwidth}{!}{
\begin{tabular}{{l}*{23}{c}}
\toprule
 & \multicolumn{23}{c}{Split 2 mAP} \\
\cmidrule(lr){2-24}
&  \multicolumn{6}{c}{Novel} & \multicolumn{16}{c}{Base} & \multirow{2}{*}{\textbf{mAP}}\\
\cmidrule(lr){2-7}  \cmidrule(lr){8-23}
Method & aero & bottle & cow & horse & sofa & \textbf{mAP} & bike & bird & boat & bus & car & cat & chair & table & dog & mbike & person & plant & sheep & train & tv & \textbf{mAP} \\
\midrule
LSTD (YOLO)-full~\cite{hao:2018:lstd} & 3.0 & 1.5 & 13.9 & 0.6 & 0.0 &  3.8 & 77.2 & 69.0 & 58.2 & 77.6 & 77.1 & 86.3 & 45.6 & 70.2 & 79.1 & 76.3 & 72.7 & 40.3 & 59.4 & 81.1 & 74.4 & 69.6 & 53.2\\
YOLO-Low-Shot~\cite{bingyi:2019:reweight} & 28.6 & 0.9 & 27.6 & 0.0 & 19.5 &  15.3 & 75.8 & 67.4 & 52.4 & 74.8 & 76.6 & 82.5 & 44.5 & 66.0 & 79.4 & 76.2 & 68.2 & 42.3 & 53.8 & 76.6 & 71.0 & 67.2 & 54.2\\
Meta R-CNN~\cite{xiaopeng:2019:metarcnn} & 12.4 & 0.1 & 44.4 & 50.1 & 0.1 &  19.4 & 69.4 & 57.7 & 38.8 & 67.1 & 71.9 & 71.7 & 24.8 & 52.6 & 60.8 & 60.7 & 72.2 & 19.7 & 59.2 & 63.0 & 55.5 & 56.3 & 47.1\\
\midrule
RFB-ft-full & 6.9 & 0.5 & 11.5 & 9.1 & 12.5 &  8.1 & 77.1 & 59.0 & 51.4 & 72.3 & 77.7 & 73.1 & 37.3 & 65.5 & 62.8 & 73.6 & 47.3 & 38.1 & 50.1 & 80.4 & 65.0 & 62.0 & 48.6\\
AttFDNet (BU'+TD) & 32.2 & 9.1 & 27.2 & 13.5 & 22.4 &  20.9 & 82.4 & 63.7 & 58.9 & 81.9 & 84.0 & 82.2 &  48.0 & 70.4 & 74.3 & 79.9 & 73.4 & 45.9 & 57.4 & 83.0 & 72.0 & 70.5 & 58.1\\
AttFDNet (BU+TD) & 36.0 & 9.2 & 29.8 & 9.1 & 19.4 & 20.7 & 82.6 & 65.1 & 58.0 & 81.4 & 84.1 & 84.0 &  46.4 & 68.0 & 73.4 & 79.3 & 73.0 & 45.7 & 55.7 & 82.6 & 71.1 & 70.0 & 57.7\\
\midrule
\midrule
 & \multicolumn{23}{c}{Split 3 mAP} \\
\cmidrule(lr){2-24}
&  \multicolumn{6}{c}{Novel} & \multicolumn{16}{c}{Base} & \multirow{2}{*}{\textbf{mAP}}\\
\cmidrule(lr){2-7}  \cmidrule(lr){8-23}
Method & boat & cat & mbike & sheep & sofa & \textbf{mAP} & aero & bike & bird & bottle & bus & car & chair & cow & table & dog & horse & person & plant & train & tv & \textbf{mAP} \\
\midrule
LSTD (YOLO)-full~\cite{hao:2018:lstd} & 0.2 & 27.3 & 0.1 & 15.0 & 0.2 & 8.5 & 77.4 & 73.3 & 69.5 & 44.8 & 78.5 & 79.2 & 43.0 & 69.2 & 66.4 & 71.9 & 82.0 & 72.3 & 39.8 & 84.5 & 69.3 & 68.1 & 53.2\\
YOLO-Low-Shot~\cite{bingyi:2019:reweight} & 6.3 & 47.1 &28.4 & 28.1 & 18.2 & 25.6 & 75.8 & 73.0 & 66.4 & 40.0 & 77.8 & 77.6 & 43.1 & 62.6 & 58.5 & 71.0 & 78.9 & 67.0 & 41.2 & 77.0 & 70.0 & 65.3 & 55.4\\
Meta R-CNN~\cite{xiaopeng:2019:metarcnn} & 10.6 & 24.0 & 36.2 & 19.2 & 0.8 & 18.2 & 67.1 & 71.3 & 68.8 & 47.7 & 77.7 & 77.2 & 42.4 & 58.9 & 63.2 & 61.9 & 80.4 & 77.0 & 29.2 & 73.3 & 68.6 & 64.3 & 52.8\\
\midrule
RFB-ft-full & 0.1 & 21.4 & 6.1 & 5.0 & 20.3 & 10.8 & 76.6 & 74.7 & 62.6 & 39.7 & 76.2 & 77.4 & 31.6 & 62.6 & 60.2 & 57.9 & 78.4 & 54.0 & 36.4 & 81.6 & 65.4 & 62.4 & 49.5\\
AttFDNet (BU'+TD) &14.1 & 41.3 & 38.9 & 21.9 & 28.0 & 28.8 & 78.2 & 81.9 & 71.7 & 47.7& 82.0 & 84.8 & 44.4 & 67.6 & 67.3 & 63.6 & 84.1 & 75.3 & 44.5 & 83.3 & 71.2 & 69.8 & 59.6\\
AttFDNet (BU+TD) &15.2 & 38.9 & 46.2 & 22.8 & 22.6 & 29.1 & 74.1 & 76.6 & 65.6 & 41.6 & 78.4 & 81.7 & 38.3 & 61.2 & 68.2 & 55.7 & 80.5 & 74.3 & 40.7 & 82.0 & 69.9 & 65.9 & 56.7\\
\bottomrule
\end{tabular}
}
\end{center}
\end{table*}

Table~\ref{table:detail-compare} presents the detailed evaluation results.
In general, the proposed AttFDNet (BU+TD) or AttFDNet (BU'+TD) achieves a high overall mean average precision (mAP, averaged over all base and novel categories) on both splits. It is noteworthy that all the few-shot object detectors trained with the novel categories have decreased APs on the base categories, compared with their base detectors.
However, the proposed AttFDNet (BU+TD) considerably outperforms the state-of-the-art and RFB-ft-full on base categories.
Such results suggest that the bottom-up attention mechanism and hybrid training strategy effectively keep the performance of base categories from catastrophic forgetting. They also suggest that the hybrid training strategy alleviates the negative impacts from the random selection of the support set. So without sacrificing the detection performance on base categories, the proposed AttFDNet (BU+TD) and AttFDNet (BU'+TD) effectively learn novel category features to improve its performance on the novel categories.

\noindent\textbf{MS COCO.} We evaluate $10$-shot and $30$-shot scenarios on the MS~COCO~\cite{tsung:2014:mscoco} benchmark with the standard metrics AP, AP$_{50}$, and AP$_{75}$. The evaluation results on the novel categories are presented in Table~\ref{table:mscoco}. Our AttFDNet (BU+TD) is consistently better than RFB-ft-full (\eg $+41.8\%$ AP$_{75}$ for $10$-shot), suggesting that few-shot object detectors can gain benefits from the introduced top-down and bottom-up attention. Our proposed AttFDNet (BU+TD) outperforms LSTD (YOLO)-full~\cite{hao:2018:lstd} (\eg $+561.9\%$ AP$_{75}$ for $10$-shot), YOLO-Low-Shot~\cite{bingyi:2019:reweight}(\eg $+202.2\%$ AP$_{75}$ for $10$-shot) and MetaDet~\cite{yuxiong:2019:metadet} (\eg $+127.9\%$ AP$_{75}$ for $10$-shot) in all the cases. It also outperforms the Meta R-CNN~\cite{xiaopeng:2019:metarcnn} in $5$ out of $6$ cases (\eg $+110.6\%$ AP$_{75}$ for $10$-shot). These significant performance improvements demonstrate that our AttFDNet (BU+TD) can achieve better bounding box regression and object category classification. Our AttFDNet (BU'+TD) also performs better than
the baseline and state-of-the-art approaches.

\noindent\textbf{MS~COCO to PASCAL~VOC.} We evaluate our method with 10-shot data of each categories from PASCAL. The mAP of AttFDNet (BU'+TD) is $33.9\%$, while the proposed AttFDNet (BU+TD) achieves $40.3\%$ mAP, compared with LSTD (YOLO)-full~\cite{hao:2018:lstd} $29.0\%$, YOLO-Low-Shot~\cite{bingyi:2019:reweight} $32.3\%$ mAP and Meta R-CNN~\cite{xiaopeng:2019:metarcnn} $37.4\%$ mAP. The performance of the baseline (RFB-ft-full) is $28.9\%$ mAP. It suggests that bottom-up attention can also provide extra information to boost the object detection performance across different benchmarks.

\begin{table*}[t]
\begin{center}
\caption{Few-shot detection performance on MS~COCO 2017 validation set for novel categories. We compare the state-of-the-art methods with baseline (RFB-ft-full) and AttFDNet (BU'+TD: bottom-up (BMS) and top-down, BU+TD: bottom-up (SAM) and top-down) under $10$-shot and $30$-shot scenarios of novel categories.
}\label{table:mscoco}
\scriptsize
\begin{tabular}{{l}*{6}{c}}
\toprule
 & \multicolumn{3}{c}{10-shot} & \multicolumn{3}{c}{30-shot}\\
\cmidrule(lr){2-4} \cmidrule(lr){5-7}
Method & AP & AP$_{50}$ & AP$_{75}$ & AP & AP$_{50}$ & AP$_{75}$\\
\midrule
LSTD (YOLO)-full~\cite{hao:2018:lstd} & 3.2 & 8.1 & 2.1 & 6.7 & 15.8 & 5.1 \\
YOLO-Low-Shot~\cite{bingyi:2019:reweight} & 5.6 & 12.3 & 4.6 & 9.1 & 19.0 & 7.6 \\
Meta R-CNN~\cite{xiaopeng:2019:metarcnn} & 8.7 & 19.1 & 6.6 & 12.4 & 25.3 & 10.8 \\
MetaDet~\cite{yuxiong:2019:metadet} & 7.1 & 14.6 & 6.1 & 11.3 & 21.7 & 8.1 \\
\midrule
RFB-ft-full  & 9.2 & 13.9 & 9.8 & 12.0 & 18.6 & 13.0 \\
AttFDNet (BU'+TD)  &  9.5 & 15.4 &10.0 & 12.0 & 19.8 & 12.1 \\
AttFDNet (BU+TD) & 12.9 & 19.5 & 13.9 & 16.3 & 24.6 & 17.3\\
\bottomrule
\end{tabular}
\end{center}
\end{table*}

\subsection{Qualitative Analysis}\label{subsec:qualitative}

\begin{figure*}[t]
\centering
\includegraphics[width=1\textwidth]{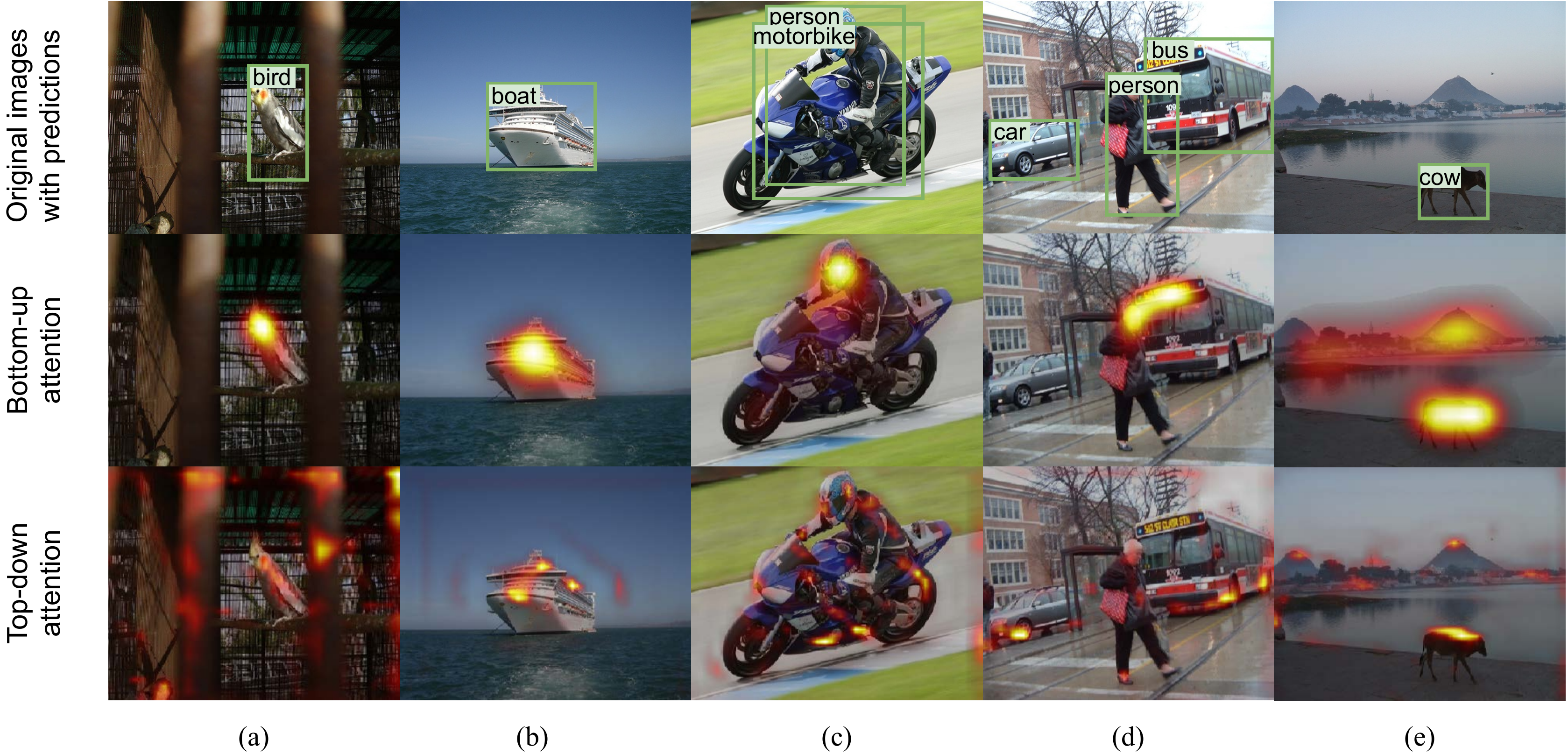}
\caption{Top-down attention and bottom-up attention are complementary. From left to right: images with detected objects annotated in green from the results of our AttFDNet (BU+TD), images overlaid with bottom-up attention maps and top-down attention maps, respectively.}
\label{fig:complementary}
\end{figure*}

Next, we compare qualitative results between the AttFDNets with and without bottom-up attention. As shown in \fig~\ref{fig:vis-rlts}, without the bottom-up attention, the baseline detector AttFDNet (TD) either fails to detect the objects or easily overfit the support set. Differently, with an off-the-shelf saliency model (\ie~SAM), our AttFDNet (BU+TD) can significantly improve the object detection performance on base and novel categories. Bottom-up attention can (a) help object detectors remember previously acquired knowledge, (b) classify object categories more accurately, (c) avoid missed classification, (d) reduce ambiguous results, and (e) improve the precision of bounding box localization. These improvements are mostly due to the complementary nature of bottom-up attention and top-down attention. A comparative analysis on the PASCAL VOC 2007 test set shows week correlations between the bottom-up and top-down attention maps (Pearson's $r = 0.208$, Spearman's $\rho = 0.205$), suggesting that the two attention mechanisms highlight different regions of interest.
\fig~\ref{fig:complementary} further demonstrate the complementary nature of the two attention mechanisms. We summarize three typical scenarios where the two attention mechanisms complement each other:

\begin{figure*}[t]
\centering
\includegraphics[width=1\textwidth]{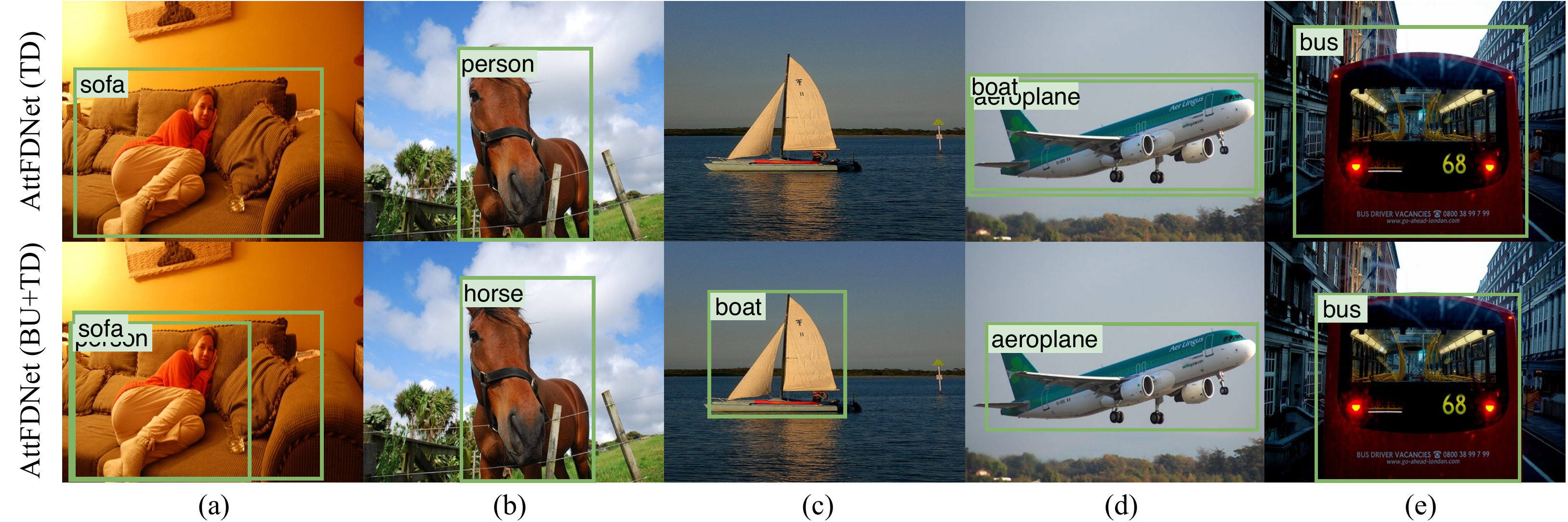}
\caption{Qualitative results of $2$-shot object detection. Detected objects are annotated in green.}
\label{fig:vis-rlts}
\end{figure*}

\noindent\textbf{Scenario I:} Bottom-up attention can successfully capture the objects of interest. In this scenario, compared with top-down attention, bottom-up attention plays an important role by placing a high priority on the salient regions related to the object to detect. As shown in the \fig~\ref{fig:complementary}a-b, the bottom-up attention itself can result in correct object detection (\ie~\emph{bird} and \emph{boat}), while top-down attention either fails to have a clearly focused region or attends to similar regions as bottom-up attention.

\noindent\textbf{Scenario II:} Bottom-up attention may sometimes focus only on a part of an object (\eg~\fig~\ref{fig:complementary}c), or highlight some but not all objects of interests  (\eg~\fig~\ref{fig:complementary}d). Under such circumstances, top-down attention can focus on different regions/objects that bottom-up attention misses. For example, in \fig~\ref{fig:complementary}c, bottom-up attention only highlights the upper part of the \emph{person} and a small part of the \emph{mbike}. Therefore, the top-down attention plays an important role in detecting the entire bounding box of the \emph{person} and the \emph{mbike}. In addition, in \fig~\ref{fig:complementary}d, we can observe that bottom-up attention highlights the \emph{person} and \emph{bus}, but not the \emph{car}. The top-down attention, on the other hand, is directed to the \emph{car} and the \emph{bus} but misses the \emph{person}. As a result, the object detector jointly considers the different regions highlighted by the bottom-up and top-down attention maps, to detect all the three objects.

\noindent\textbf{Scenario III:} There are also cases where top-down attention can play a more important role in detecting objects of interest.
For example, as shown in \fig~\ref{fig:complementary}e, due to the complexity of the scene and the relatively small object region (\ie~the \emph{cow} walking on the bank), the bottom-up attention not only highlights the \emph{cow}, but also the background. In this case, the object detector relies on top-down attention to exclude a large area of irrelevant regions and pay more attention to the more related regions.

\begin{figure*}[htbp]
\centering
\includegraphics[width=1\textwidth]{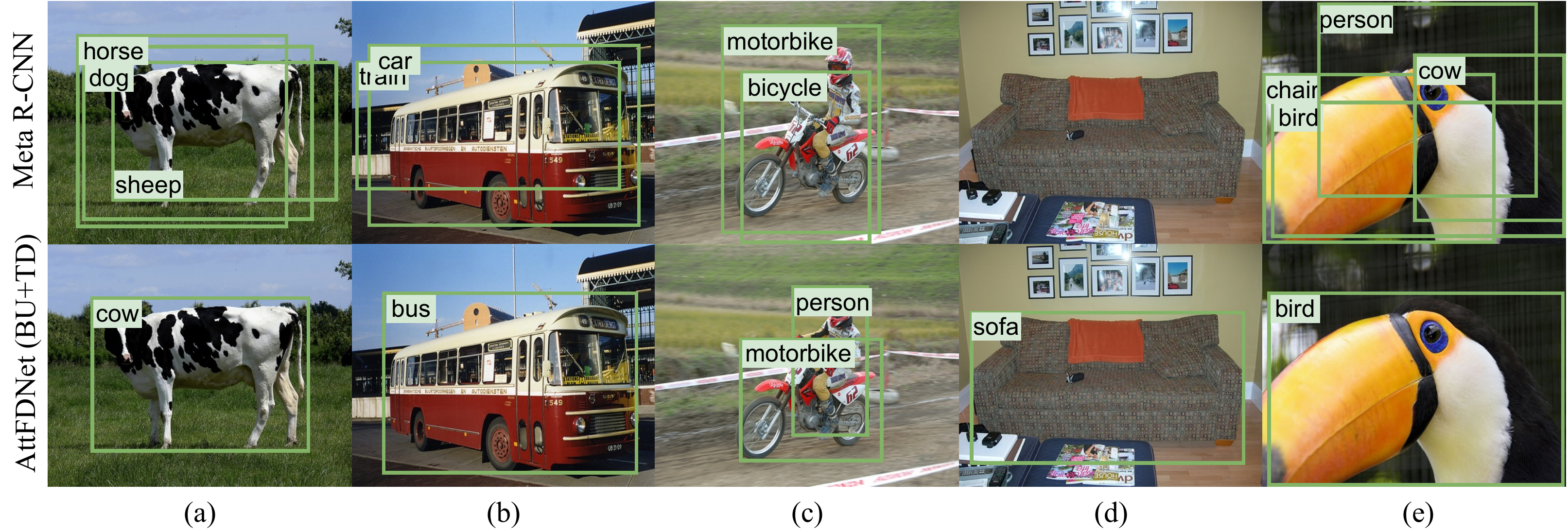}
\caption{Qualitative examples of $2$-shot object detection results on PASCAL VOC. We compare our proposed AttFDNet (BU+TD) with Meta R-CNN~\cite{xiaopeng:2019:metarcnn}. Detected bounding boxes are shown in green.}
\label{fig:sup-vis-rlts}
\end{figure*}

We also demonstrate the effectiveness of our proposed AttFDNet (BU+TD) by comparing the qualitative results between AttFDNet (BU+TD) and one of the state-of-the-art methods Meta R-CNN~\cite{xiaopeng:2019:metarcnn}.
As shown in Fig.~\ref{fig:sup-vis-rlts}, AttFDNet (BU+TD) predicts the object categories more accurately and positions the bounding boxes more precisely. Specifically, we observe four typical benefits from the incorporation of bottom-up attention:
\begin{enumerate*}
\item[(1)] Object categories are classified more accurately (\eg~\textit{cow} in \fig~\ref{fig:sup-vis-rlts}a and \textit{bus} in \fig~\ref{fig:sup-vis-rlts}b).
\item[(2)] Bounding box localization is more precise (\eg~\textit{motorbike} in \fig~\ref{fig:sup-vis-rlts}c).
\item[(3)] The detector better remembers previously acquired knowledge (\eg~\textit{sofa} in \fig~\ref{fig:sup-vis-rlts}d);
\item[(4)] The number of ambiguous results is reduced (\eg~\textit{bird} in \fig~\ref{fig:sup-vis-rlts}e);
\end{enumerate*}
These observations suggest that bottom-up attention can highlight the globally salient features so that the detector can better detect and recognize such regions, leading to the improved performances.

In sum, the qualitative examples suggest that bottom-up attention and top-down attention localize different regions of interest in a complementary manner, which jointly improves the detection performance.

\subsection{Ablation Studies}

We present results of comprehensive ablation studies to analyze the effects of various components.  All ablation studies are conducted on the PASCAL VOC 2007 test set for the $2$-shot scenario on the second and third splits.

\noindent\textbf{Effects of the backbone.} First, we compare
the performances of two backbone networks (\ie~VGG~\cite{liang:2018:deeplabvgg,kares:2015:vgg} and ResNet-101~\cite{kaiming:2016:resnet}) on detecting base and novel object categories. Table~\ref{table:ablation-backbone} shows that VGG outperforms the ResNet-101 model on both base and novel categories.
The bottom-up attention consistently improves the performance of few-shot object detectors despite the different backbones used.

For simplification, we use VGG backbone in the all the other experiments since such one-stage detectors SSD~\cite{wei:2016:ssd} and RFB~\cite{songtao:2018:rfb} also use VGG backbone.

\noindent\textbf{Effects of the hyper-parameters.}  With a grid search, we optimize the hyper-parameters $\beta$ and $\eta$ that control the weights of the proposed object-concentration loss and background-concentration loss. Table~\ref{table:ablation-parameter} shows that the introduction of concentration loss terms can improve the performance of few-shot detection on the novel categories. It is also noteworthy that improving the performance of the novel categories could decrease the performance on the base categories. Based on it, we chose $\beta=2$ and $\eta=0.4$ as the hyper-parameters for all the experiments, as they result in the highest overall mAP.  Furthermore, we analyze the 2-shot scenario for split 2 in the test set. The cosine similarity between the features of the ground truth prior anchors and their corresponding weights of the final fully-connected layer is increased from $0.452$ to $0.573$ compared with the elimination of the object-concentration loss, which demonstrates that object-concentration loss can improve the intra-class agreement of features.

We further conduct ablation studies on the selection of the hyper-parameter $\epsilon$ that controls the smoothness of the bottom-up attention when it is integrated into the model. 
A larger $\epsilon$ indicates a more smooth integration and hence assigns similar weights for salient/non-salient regions of the feature map.
When $\epsilon$ is too large, the proposed AttFDNet (BU+TD) would degrade to AttFDNet (TD), since the model cannot fully utilize the information from the bottom-up attention to boost the few-shot object detection in this case.
A smaller $\epsilon$ indicates a sharper integration, which means that we assign more weights to the salient regions than the non-salient regions.
When $\epsilon$ is too small,
we would assign an approximately zero weight (even when $\epsilon=1$, this weight would become $0$) for non-salient regions to generate the feature map, which will prevent the detectors from detecting the object located in the non-salient regions.
According to Table~\ref{table:ablation-parameter}, we choose $\epsilon=e$ for all the experiments, since it obtains the best overall mAP scores. When $\epsilon=e$, the non-salient regions for the feature map would be unchanged, since $\ln(e+g(\mathbf{x};\varphi))=1$ given $g(\mathbf{x};\varphi)=0$. Hence, it would not hurt the detection performance for the non-salient region and can provide enough information from the salient regions to boost the final performance.

Lastly, we also optimize the hyper-parameter $\gamma$ that controls the weights of the proposed distillation. Table~\ref{table:ablation-parameter} shows the performance among different parameter settings. On the one hand, a large $\gamma$ indicates more attention would be paid on the performance of the base categories and hence ignores the performance of the novel categories. When $\gamma$ is too large (\eg $\gamma=1.0$), the proposed AttFDNet (BU+TD) would focus on the performance of the base categories, while the performance of novel categories would degrade as indicated in Table~\ref{table:ablation-parameter}. On the other hand, a small $\gamma$ means that the network would focus on the performance of the novel categories. Due to the few-shot object detection scenario, it would be easy to overfit the novel training dataset. More severely, it would also be harmful to the learned feature representation of the different aspects of the object and hence lead to the degradation of the overall performance. When $\gamma$ is too small (\eg $\gamma=0.0$), the proposed AttFDNet (BU+TD) would focus on the performance of the novel categories. It leads to a catastrophic forgetting on the base categories and losses the learned feature representation as indicated in Table~\ref{table:ablation-parameter}.

\begin{table}[t]
\begin{center}
\caption{The ablation study (TD: top-down only, BU+TD: bottom-up and top-down) of network backbones VGG~\cite{liang:2018:deeplabvgg,kares:2015:vgg} and ResNet-101~\cite{kaiming:2016:resnet} for the 2-shot scenario.}
\label{table:ablation-backbone}
\resizebox{1\linewidth}{!}{
\begin{tabular}{cccccc}
\toprule
\# Split & Backbone & Attention & Base mAP & Novel mAP& All mAP\\
\midrule
\multirow{4}{*}{2} &  \multirow{2}{*}{VGG} & TD & 69.2 & 17.6 & 56.3 \\
&  & BU+TD & \textbf{70.0} & \textbf{20.7} & \textbf{57.7} \\
\cmidrule(lr){2-6}
& \multirow{2}{*}{ResNet-101} & TD & \textbf{67.2} & 18.4 & 55.0 \\
&   & BU+TD & 66.5 & \textbf{21.2} & \textbf{55.2} \\
\midrule
\multirow{4}{*}{3}  &  \multirow{2}{*}{VGG} & TD & 65.7 & 26.1 & 55.8 \\
&  & BU+TD & \textbf{65.9} & \textbf{29.1} & \textbf{56.7} \\
\cmidrule(lr){2-6}
& \multirow{2}{*}{ResNet-101} & TD & 62.0 & 26.3 & 53.1 \\
&  & BU+TD & \textbf{63.4} & \textbf{28.6} & \textbf{54.7} \\
 \bottomrule

\end{tabular}
}
\end{center}
\end{table}

\begin{table}[t]
\begin{center}
\caption{The ablation study of the hyper-parameters $\beta$, $\eta$ $\epsilon$, and $\gamma$ for the 2-shot scenario.}
\label{table:ablation-parameter}
\scriptsize
\resizebox{1\linewidth}{!}{
\begin{tabular}{ccccc}
\toprule
\# Split & Hyper-parameters & Base mAP & Novel mAP& All mAP\\
\midrule
\multirow{12}{*}{2} &   $\beta=0, \eta = 0.4, \epsilon=e, \gamma = 0.5$ & 67.6 & 19.2 & 55.5 \\
 &  $\beta=2, \eta = 0.4, \epsilon=e, \gamma = 0.5$ & 70.0 & 20.7 & \textbf{57.7} \\
 &  $\beta=5, \eta = 0.4, \epsilon=e, \gamma = 0.5$ & 64.2 & 20.4 & 53.2 \\
\cmidrule(lr){2-5}
 & $\beta=2, \eta=0.0, \epsilon=e, \gamma = 0.5$ & 69.8 & 20.7 & 57.5 \\
 & $\beta=2, \eta=0.4, \epsilon=e, \gamma = 0.5$ & 70.0 & 20.7 & \textbf{57.7} \\
 & $\beta=2, \eta=1.0, \epsilon=e, \gamma = 0.5$ & 64.8 & 17.9 & 53.1 \\
\cmidrule(lr){2-5}
 & $\beta=2, \eta=0.4, \epsilon=1, \gamma = 0.5$ & 62.7 & 19.8 & 51.9 \\
 & $\beta=2, \eta=0.4, \epsilon=e, \gamma = 0.5$ & 70.0 & 20.7 & \textbf{57.7} \\
 & $\beta=2, \eta=0.4, \epsilon=5, \gamma = 0.5$ & 66.1 & 19.5 & 54.4 \\
\cmidrule(lr){2-5}
 & $\beta=2, \eta=0.4, \epsilon=e, \gamma = 0.0$ & 52.0 & 18.0 & 43.5 \\
 & $\beta=2, \eta=0.4, \epsilon=e, \gamma = 0.5$ & 70.0 & 20.7 & \textbf{57.7} \\
 & $\beta=2, \eta=0.4, \epsilon=e, \gamma = 1.0$ & 67.4 & 18.5 & 55.2 \\
\midrule
\multirow{12}{*}{3} & $\beta=0, \eta = 0.4, \epsilon=e, \gamma = 0.5$ & 66.2 & 26.9 & 56.4 \\
 & $\beta=2, \eta = 0.4, \epsilon=e, \gamma = 0.5$ & 65.9 & 29.1 & \textbf{56.7} \\
 & $\beta=5, \eta = 0.4, \epsilon=e, \gamma = 0.5$ & 67.4 & 18.5 & 55.2 \\
\cmidrule(lr){2-5}
 & $\beta=2, \eta=0.0, \epsilon=e, \gamma = 0.5$ &65.9 & 27.5 & 56.3 \\
 & $\beta=2, \eta=0.4, \epsilon=e, \gamma = 0.5$ & 65.9 & 29.1 & \textbf{56.7} \\
 & $\beta=2, \eta=1.0, \epsilon=e, \gamma = 0.5$ & 64.9 & 27.8 & 55.6 \\
\cmidrule(lr){2-5}
 & $\beta=2, \eta=0.4, \epsilon=1, \gamma = 0.5$ & 62.1 & 28.6 & 53.7 \\
 & $\beta=2, \eta=0.4, \epsilon=e, \gamma = 0.5$ & 65.9 & 29.1 & \textbf{56.7} \\
 & $\beta=2, \eta=0.4, \epsilon=5, \gamma = 0.5$ & 65.8 & 25.9 & 55.8 \\
\cmidrule(lr){2-5}
 & $\beta=2, \eta=0.4, \epsilon=e, \gamma = 0.0$ & 52.0 & 24.2 & 45.1 \\
 & $\beta=2, \eta=0.4, \epsilon=e, \gamma = 0.5$ & 65.9 & 29.1 & \textbf{56.7} \\
 & $\beta=2, \eta=0.4, \epsilon=e, \gamma = 1.0$ & 66.7 & 25.6 & 56.4 \\
\bottomrule
\end{tabular}
}
\end{center}
\end{table}

\begin{table}[t]
\begin{center}
\caption{The ablation study of the different modules for the 2-shot scenario.}
\label{table:ablation-parameter}
\scriptsize
\begin{tabular}{cl|ccc}
\toprule
\# Split & Method & Base mAP & Novel mAP& All mAP\\
\midrule
  \multirow{6}{*}{2} & RFB-ft-full & 62.0 & 8.1 & 48.6\\
   & AttFDNet (TD) & 69.2 & 17.6 & 56.3 \\
   & AttFDNet (BU'+TD) & \textbf{70.5} & \textbf{20.9} & \textbf{58.1} \\
   & AttFDNet (BU+TD) & 70.0 & 20.7 & 57.7 \\
   & AttFDNet (BU+TD) w/o distillation  & 52.0 & 18.0 & 43.5 \\
   & AttFDNet (BU+TD) w/o bk  & 69.8 & 20.7 & 57.5 \\
   & AttFDNet (BU+TD) w/o (bk+obj) & 67.5 & 19.8 & 55.4  \\
\midrule
  \multirow{6}{*}{3} & RFB-ft-full & 62.4 & 10.8 & 49.5\\
   & AttFDNet (TD) & 65.7 & 26.1 & 55.8 \\
   & AttFDNet (BU'+TD) & \textbf{69.8} & 28.8 & \textbf{59.6} \\
   & AttFDNet (BU+TD) & 65.9 & \textbf{29.1} & 56.7 \\
   & AttFDNet (BU+TD) w/o distillation  & 52.0 & 24.2 & 45.1 \\
   & AttFDNet (BU+TD) w/o bk  & 65.9 & 27.5 & 56.3 \\
   & AttFDNet (BU+TD) w/o (bk+obj) & 66.0 & 26.6 & 56.1  \\
\bottomrule
\end{tabular}
\end{center}
\end{table}

\noindent\textbf{Effects of the different modules.} We investigate the effectiveness of the proposed modules. Compared AttFDNet (TD) with AttFDNet (BU'+TD) and AttFDNet (BU+TD), we observe that the bottom-up attention can provide extra information to improve the performance. The improvement of the use of SAM and BMS also demonstrates the robustness of our proposed bottom-up attention.
We can also observe that all AttFDNet models (TD, BU'+TD and BU+TD) significantly outperform the baseline (RFB-ft-full).
Next, we discuss the three loss terms introduced in Equation~\eqref{equ:novel-detection-loss}.
AttFDNet (BU+TD) w/o distillation represents the ablation of distillation module from our AttFDNet (BU+TD), while AttFDNet (BU+TD) w/o bk means the ablation of background concentration loss from our AttFDNet (BU+TD). Furthermore, AttFDNet (BU+TD) w/o (bk+obj) is the ablation of background concentration loss and object concentration loss from our AttFDNet (BU+TD).
Without the distillation loss, the performance of the base mAP degrades significantly, which means that distillation loss can play an important role in remember the previous detection tasks. We can also observe that the concentration loss can still help to boost the performance of base and novel categories.

\section{Conclusion}
\label{sec:concluision}
In this paper, we have introduced a novel attentive few-shot object
detector that incorporates bottom-up and top-down attention for detecting novel object categories with extremely few training samples. Learning from human attention data, bottom-up attention provides prior knowledge about salient regions and plays a different role from the top-down attention learned from the object annotations. To address specific challenges in few-shot object detection, we also propose a hybrid few-shot learning strategy and two concentration loss terms. The proposed detector achieve state-of-the-art performances in extremely few-shot scenarios, demonstrating the significant and complementary roles of the two attention mechanisms. Future efforts will be focused on the exploration of different attention fusion methods to make the best use of bottom-up and top-down attention in few-shot object detection and other vision tasks.

\ifCLASSOPTIONcaptionsoff
  \newpage
\fi



%


\bibliographystyle{IEEEtran}
\bibliography{egbib}

\begin{thebibliography}{10}
\providecommand{\url}[1]{#1}
\csname url@samestyle\endcsname
\providecommand{\newblock}{\relax}
\providecommand{\bibinfo}[2]{#2}
\providecommand{\BIBentrySTDinterwordspacing}{\spaceskip=0pt\relax}
\providecommand{\BIBentryALTinterwordstretchfactor}{4}
\providecommand{\BIBentryALTinterwordspacing}{\spaceskip=\fontdimen2\font plus
\BIBentryALTinterwordstretchfactor\fontdimen3\font minus
  \fontdimen4\font\relax}
\providecommand{\BIBforeignlanguage}[2]{{%
\expandafter\ifx\csname l@#1\endcsname\relax
\typeout{** WARNING: IEEEtran.bst: No hyphenation pattern has been}%
\typeout{** loaded for the language `#1'. Using the pattern for}%
\typeout{** the default language instead.}%
\else
\language=\csname l@#1\endcsname
\fi
#2}}
\providecommand{\BIBdecl}{\relax}
\BIBdecl

\bibitem{hao:2018:lstd}
H.~Chen, Y.~Wang, G.~Wang, and Y.~Qiao, ``Lstd: A low-shot transfer detector
  for object detection,'' \emph{Association for the Advancement of Artificial
  Intelligence (AAAI)}, 2018.

\bibitem{ze:2020:ctfsod}
Z.~Yang, Y.~Wang, X.~Chen, and Y.~Qiao, ``Context-transformer: Tackling object
  confusion for few-shot detection.'' \emph{Association for the Advancement of
  Artificial Intelligence (AAAI)}, 2020.

\bibitem{xin:2020:fsod}
X.~Wang, T.~E. Huang, T.~Darrell, J.~E. Gonzalez, and F.~Yu, ``Frustratingly
  simple few-shot object detection,'' \emph{International Conference on Machine
  Learning (ICML)}, 2020.

\bibitem{bingyi:2019:reweight}
B.~Kang, Z.~Liu, X.~Wang, F.~Yu, J.~Feng, and T.~Darrell, ``Few-shot object
  detection via feature reweighting,'' \emph{IEEE International Conference on
  Computer Vision (ICCV)}, 2019.

\bibitem{xiaopeng:2019:metarcnn}
X.~Yan, Z.~Chen, A.~Xu, X.~Wang, X.~Liang, and L.~Lin, ``Meta {R-CNN} : Towards
  general solver for instance-level low-shot learning,'' \emph{IEEE
  International Conference on Computer Vision (ICCV)}, 2019.

\bibitem{yuxiong:2019:metadet}
Y.-X. Wang, D.~Ramanan, and M.~Hebert, ``Meta-learning to detect rare
  objects,'' \emph{IEEE International Conference on Computer Vision (ICCV)},
  2019.

\bibitem{qi:2019:fsod}
Q.~Fan, W.~Zhuo, and Y.-W. Tai, ``Few-shot object detection with attention-rpn
  and multi-relation detector,'' \emph{CoRR, abs/1908.01998}, 2019.

\bibitem{yiting:2020:lscn}
Y.~Li, Y.~Cheng, L.~Liu, S.~Tian, H.~Zhu, C.~Xiang, P.~Vadakkepat, C.~Teo, and
  T.~Lee, ``Low-shot object detection via classification refinement,''
  \emph{CoRR, abs/2005.02641}, 2020.

\bibitem{juanmanuel:2020:once}
J.-M. Perez-Rua, X.~Zhu, T.~Hospedales, and T.~Xiang, ``Incremental few-shot
  object detection,'' \emph{IEEE Conference on Computer Vision and Pattern
  Recognition (CVPR)}, 2020.

\bibitem{ting:2019:coae}
T.-I. Hsieh, Y.-C. Lo, H.-T. Chen, and T.-L. Liu, ``One-shot object detection
  with co-attention and co-excitation,'' \emph{Conference on Neural Information
  Processing Systems (NeurIPS)}, 2019.

\bibitem{shafin:2020:asd}
S.~Rahman, S.~Khan, N.~Barnes, and F.~S. Khan, ``Any-shot object detection,''
  \emph{CoRR, abs/2003.07003}, 2020.

\bibitem{siddhesh:2020:waod}
S.~Khandelwal, R.~Goyal, and L.~Sigal, ``Weakly-supervised any-shot object
  detection,'' \emph{CoRR, abs/2006.07502}, 2020.

\bibitem{tam:2018:attentivesystem}
T.~V. Nguyen, Q.~Zhao, and S.~Yan, ``Attentive systems: A survey.''
  \emph{International Journal of Computer Vision (IJCV)}, 2018.

\bibitem{milica:2012:saliencybias}
M.~Milosavljevic, V.~Navalpakkam, C.~Koch, and A.~Rangel, ``Relative visual
  saliency differences induce sizable bias in consumer choice.'' \emph{Journal
  of Consumer Psychology (JCP)}, 2012.

\bibitem{lori:2012:onsaliency}
L.~McCay-Peet, M.~Lalmas, and V.~Navalpakkam, ``On saliency, affect and focused
  attention.'' \emph{Proceedings of the SIGCHI Conference on Human Factors in
  Computing Systems (SIGCHI)}, 2012.

\bibitem{chinkai:2010:visionnavigation}
C.-K. Chang, C.~Siagian, and L.~Itti, ``Mobile robot vision navigation \&
  localization using gist and saliency.'' \emph{IEEE/RSJ International
  Conference on Intelligent Robots and Systems (IROS)}, 2010.

\bibitem{chinkai:2011:visionnavigation}
------, ``Mobile robot vision navigation and obstacle avoidance based on gist
  and saliency algorithms.'' \emph{Journal of Vision (JoV)}, 2011.

\bibitem{dapeng:2015:saliencydetection}
D.~Tao, J.~Cheng, M.~Song, and X.~Lin, ``Manifold ranking-based matrix
  factorization for saliency detection.'' \emph{IEEE Transactions on Neural
  Networks and Learning Systems (TNNLS)}, 2015.

\bibitem{tam:2019:salientod}
T.~V. Nguyen, K.~Nguyen, and T.-T. Do, ``Semantic prior analysis for salient
  object detection.'' \emph{IEEE Transactions on Image Processing (TIP)}, 2019.

\bibitem{tam:2017:salientod}
T.~V. Nguyen and L.~Liu, ``Salient object detection with semantic priors.''
  \emph{International Joint Conference on Artificial Intelligence (IJCAI)},
  2017.

\bibitem{mengmi:2018:anticipatelook}
M.~Zhang, K.~T. Ma, J.~H. Lim, Q.~Zhao, and J.~Feng, ``Anticipating where
  people will look using adversarial networks.'' \emph{IEEE Transactions on
  Pattern Analysis and Machine Intelligence (TPAMI)}, 2018.

\bibitem{mark:2015:voc}
M.~Everingham, S.~M.~A. Eslami, L.~V. Gool, C.~K.~I. Williams, J.~Winn, and
  A.~Zisserman, ``The pascal visual object classes challenge: A
  retrospective,'' \emph{International Journal of Computer Vision (IJCV)},
  2015.

\bibitem{mark:2010:voc}
M.~Everingham, L.~V. Gool, C.~K.~I. Williams, J.~Winn, and A.~Zisserman, ``The
  pascal visual object classes ({VOC}) challenge,'' \emph{International Journal
  of Computer Vision (IJCV)}, 2010.

\bibitem{navneet:2005:hog}
N.~Dalal and B.~Triggs, ``Histograms of oriented gradients for human
  detection,'' \emph{IEEE Conference on Computer Vision and Pattern Recognition
  (CVPR)}, 2005.

\bibitem{jasper:2013:selectivesearch}
J.~R.~R. Uijlings, K.~E.~A. Sande, T.~Gevers, and A.~W.~M. Smeulders,
  ``Selective search for object recognition,'' \emph{International Journal of
  Computer Vision (IJCV)}, 2013.

\bibitem{lawrence:2014:edgeboxes}
C.~L. Zitnick and P.~Doll{\'{a}}r, ``Edge boxes: Locating object proposals from
  edge,'' \emph{European Conference on Computer Vision (ECCV)}, 2014.

\bibitem{xingyi:2019:pointobject}
X.~Zhou, D.~Wang, and P.~Kr{\"{a}}henb{\"{u}}hl, ``Objects as points,''
  \emph{IEEE Conference on Computer Vision and Pattern Recognition (CVPR)},
  2019.

\bibitem{joseph:2016:yolo}
J.~Redmon, S.~Divvala, R.~Girshick, and A.~Farhadi, ``You only look once:
  Unified, real-time object detection,'' \emph{IEEE Conference on Computer
  Vision and Pattern Recognition (CVPR)}, 2016.

\bibitem{joseph:2016:yolo2}
J.~Redmon and A.~Farhadi, ``Yolo9000: Better, faster, stronger,'' \emph{CoRR,
  abs/1612.08242}, 2016.

\bibitem{joseph:2018:yolo3}
------, ``Yolov3: An incremental improvement,'' \emph{CoRR, abs/1804.02767},
  2018.

\bibitem{wei:2016:ssd}
W.~Liu, D.~Anguelov, D.~Erhan, C.~Szegedy, S.~Reed, C.-Y. Fu, and A.~C. Berg,
  ``Ssd: Single shot multibox detector,'' \emph{European Conference on Computer
  Vision (ECCV)}, 2016.

\bibitem{cheng-yang:2016:dssd}
C.-Y. Fu, W.~Liu, A.~Ranga, A.~Tyagi, and A.~C. Berg, ``Dssd : Deconvolutional
  single shot detector,'' \emph{IEEE Conference on Computer Vision and Pattern
  Recognition (CVPR)}, 2017.

\bibitem{songtao:2018:rfb}
S.~Liu, D.~Huang, and Y.~Wang, ``Receptive field block net for accurate and
  fast object detection,'' \emph{European Conference on Computer Vision
  (ECCV)}, 2018.

\bibitem{ross:2014:rcnn}
R.~Girshick, J.~Donahue, T.~Darrell, and J.~Malik, ``Rich feature hierarchies
  for accurate object detection and semantic segmentation,'' \emph{IEEE
  Conference on Computer Vision and Pattern Recognition (CVPR)}, 2014.

\bibitem{ross:2015:fastrcnn}
R.~Girshick, ``Fast r-cnn,'' \emph{IEEE International Conference on Computer
  Vision (ICCV)}, 2015.

\bibitem{kaiming:2017:maskrcnn}
K.~He, G.~Gkioxari, P.~Doll{\'{a}}r, and R.~Girshick, ``Mask r-cnn,''
  \emph{IEEE International Conference on Computer Vision (ICCV)}, 2017.

\bibitem{shaoqing:2017:fasterrcnn}
S.~Ren, K.~He, R.~Girshick, and J.~Sun, ``Faster r-cnn: Towards real-time
  object detection with region proposal networks,'' \emph{IEEE Transactions on
  Pattern Analysis and Machine Intelligence (TPAMI)}, 2017.

\bibitem{yaqing:2019:survey-fewshot}
Y.~Wang, Q.~Yao, J.~Kwok, and L.~M. Ni, ``Generalizing from a few examples: A
  survey on few-shot learning,'' \emph{CoRR, abs/1904.05046}, 2019.

\bibitem{chelsea:2017:maml}
S.~L. Chelsea~Finn, Pieter~Abbeel, ``Model-agnostic meta-learning for fast
  adaptation of deep networks,'' \emph{International Conference on Machine
  Learning (ICML)}, 2017.

\bibitem{weiyu:2019:fewshocls}
W.-Y. Chen, Y.-C. Liu, Z.~Kira, Y.-C.~F. Wang, and J.-B. Huang, ``A closer look
  at few-shot classification,'' \emph{International Conference on Learning
  Representations (ICLR)}, 2019.

\bibitem{li:2003:baye-oneshot}
L.~Fei-Fei, R.~Fergus, and P.~Peron, ``A bayesian approach to unsupervised
  one-shot learning of object categories,'' \emph{IEEE International Conference
  on Computer Vision (ICCV)}, 2003.

\bibitem{koch:2015:siamese}
K.~Gregory, Z.~Richard, and S.~Ruslan, ``Siamese neural networks for one-shot
  image recognition,'' \emph{International Conference on Machine Learning
  (ICML) Deep Learning Workshop}, 2015.

\bibitem{taesup:2018:bmaml}
T.~Kim, J.~Yoon, O.~Dia, S.~Kim, Y.~Bengio, and S.~Ahn, ``Bayesian
  model-agnostic meta-learning,'' \emph{Conference on Neural Information
  Processing Systems (NeurIPS)}, 2018.

\bibitem{kwonjoon:2019:metaco}
K.~Lee, S.~Maji, A.~Ravichandran, and S.~Soatto, ``Meta-learning with
  differentiable convex optimization,'' \emph{IEEE Conference on Computer
  Vision and Pattern Recognition (CVPR)}, 2019.

\bibitem{jake:2017:prototypical}
J.~Snell, K.~Swersky, and R.~S. Zemel, ``Prototypical networks for few-shot
  learning,'' \emph{Conference on Neural Information Processing Systems
  (NeurIPS)}, 2017.

\bibitem{qianru:2019:metatransfer}
Q.~Sun, Y.~Liu, T.-S. Chua, and B.~Schiele, ``Meta-transfer learning for
  few-shot learning,'' \emph{IEEE Conference on Computer Vision and Pattern
  Recognition (CVPR)}, 2019.

\bibitem{flood:2018:relation-network}
F.~Sung, Y.~Yang, L.~Zhang, T.~Xiang, P.~H. Torr, and T.~M. Hospedales,
  ``Learning to compare: Relation network for few-shot learning,'' \emph{IEEE
  Conference on Computer Vision and Pattern Recognition (CVPR)}, 2018.

\bibitem{oriol:2016:matchingnet}
O.~Vinyals, C.~Blundell, T.~Lillicrap, K.~Kavukcuoglu, and D.~Wierstra,
  ``Matching networks for one shot learning,'' \emph{Conference on Neural
  Information Processing Systems (NeurIPS)}, 2016.

\bibitem{leonid:2019:repmet}
L.~Karlinsky, J.~Shtok, S.~Harary, E.~Schwartz, A.~Aides, R.~Feris, R.~Giryes,
  and A.~M. Bronstein, ``Repmet: Representative-based metric learning for
  classification and one-shot object detection,'' \emph{IEEE Conference on
  Computer Vision and Pattern Recognition (CVPR)}, 2019.

\bibitem{xiaolong:2018:nolocal}
X.~Wang, R.~Girshick, A.~Gupta, and K.~He, ``Non-local neural networks,''
  \emph{IEEE Conference on Computer Vision and Pattern Recognition (CVPR)},
  2018.

\bibitem{jie:2018:senet}
J.~Hu, L.~Shen, S.~Albanie, G.~Sun, and E.~Wu, ``Squeeze-and-excitation
  networks,'' \emph{IEEE Conference on Computer Vision and Pattern Recognition
  (CVPR)}, 2018.

\bibitem{xianyu:2020:did}
X.~Chen, Y.~Wang, J.~Liu, and Y.~Qiao, ``Did:
  Disentangling-imprinting-distilling for continuous low-shot detection.''
  \emph{IEEE Transactions on Image Processing (TIP)}, 2020.

\bibitem{yue:2019:gcnet}
Y.~Cao, J.~Xu, S.~Lin, F.~Wei, and H.~Hu, ``Gcnet: Non-local networks meet
  squeeze-excitation networks and beyond,'' \emph{CoRR, abs/1904.11492}, 2018.

\bibitem{jan:2015:speechrecog}
J.~Chorowski, D.~Bahdanau, D.~Serdyuk, K.~Cho, and Y.~Bengio, ``Attention-based
  models for speech recognition,'' \emph{Conference on Neural Information
  Processing Systems (NeurIPS)}, 2015.

\bibitem{zhenyang:2018:videolstm}
Z.~Li, E.~Gavves, M.~Jain, and C.~G.~M. Snoek, ``Video{LSTM} convolves, attends
  and flows for action recognition,'' \emph{Computer Vision and Image
  Understanding}, 2018.

\bibitem{kelvin:2015:image-caption}
K.~Xu, J.~Ba, R.~Kiros, K.~Cho, A.~Courville, R.~Salakhutdinov, R.~Zemel, and
  Y.~Bengio, ``Show, attend and tell: Neural image caption generation with
  visual attention,'' \emph{International Conference on Machine Learning
  (ICML)}, 2015.

\bibitem{liang:2018:deeplabvgg}
L.-C. Chen, G.~Papandreou, I.~Kokkinos, K.~Murphy, and A.~L. Yuille, ``Deeplab:
  Semantic image segmentation with deep convolutional nets, atrous convolution,
  and fully connected crfs,'' \emph{IEEE Transactions on Pattern Analysis and
  Machine Intelligence (TPAMI)}, 2018.

\bibitem{kares:2015:vgg}
K.~Simonyan and A.~Zisserman, ``Very deep convolutional networks for
  large-scale image recognition,'' \emph{International Conference on Learning
  Representations (ICLR)}, 2015.

\bibitem{hang:2017:imprinting}
H.~Qi, M.~Brown, and D.~G. Lowe, ``Low-shot learning with imprinted weights,''
  \emph{IEEE Conference on Computer Vision and Pattern Recognition (CVPR)},
  2017.

\bibitem{geoffrey:2015:distill}
G.~Hinton, O.~Vinyals, and J.~Dean, ``Distilling the knowledge in a neural
  network,'' \emph{CoRR, abs/1503.02531}, 2015.

\bibitem{konstantin:2017:incremental_learning}
K.~Shmelkov, C.~Schmid, and K.~Alahari, ``Incremental learning of object
  detectors without catastrophic forgetting,'' \emph{IEEE International
  Conference on Computer Vision (ICCV)}, 2017.

\bibitem{tsung:2014:mscoco}
T.~Lin, M.~Maire, S.~J. Belongie, L.~D. Bourdev, R.~B. Girshick, J.~Hays,
  P.~Perona, D.~Ramanan, P.~Doll{\'{a}}r, and C.~L. Zitnick, ``Microsoft
  {COCO}: Common objects in context,'' \emph{European Conference on Computer
  Vision (ECCV)}, 2014.

\bibitem{marcella:2018:sam}
M.~Cornia, L.~Baraldi, G.~Serra, and R.~Cucchiara, ``Predicting human eye
  fixations via an lstm-based saliency attentive model,'' \emph{IEEE
  Transactions on Image Processing (TIP)}, 2018.

\bibitem{ming:2015:salicon}
M.~Jiang, S.~Huang, J.~Duan, and Q.~Zhao, ``{SALICON}: Saliency in context,''
  \emph{IEEE International Conference on Computer Vision (ICCV)}, 2015.

\bibitem{diederik:2015:adam}
D.~P. Kingma and J.~Ba, ``Adam: A method for stochastic optimization,''
  \emph{International Conference on Learning Representations (ICLR)}, 2015.

\bibitem{jianming:2013:bms}
J.~Zhang and S.~Sclaroff, ``Saliency detection: A boolean map approach,''
  \emph{IEEE International Conference on Computer Vision (ICCV)}, 2013.

\bibitem{kaiming:2016:resnet}
K.~He, X.~Zhang, S.~Ren, and J.~Sun, ``Deep residual learning for image
  recognition,'' \emph{IEEE Conference on Computer Vision and Pattern
  Recognition (CVPR)}, 2016.

\end{thebibliography}

%

\begin{IEEEbiography}[{\includegraphics[width=1in,height=1.25in,clip,keepaspectratio]{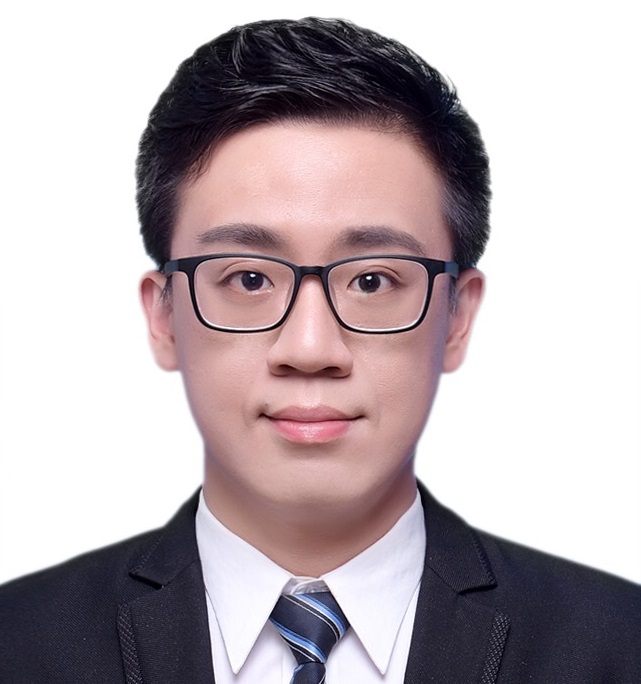}}]{Xianyu Chen} received his bachelor of engineering degree in the school of information science and technology, Sun Yat-sen University, Guangzhou, China, in 2015, and master of engineering in the school of electronics and information technology, Sun Yat-sen University, Guangzhou, China, in 2018. He is currently a Ph.D. student in the department of computer science, University of Minnesota, USA. His research interests include computer vision, pattern recognition and machine learning.
\end{IEEEbiography}

\begin{IEEEbiography}[{\includegraphics[width=1in,height=1.25in,clip,keepaspectratio]{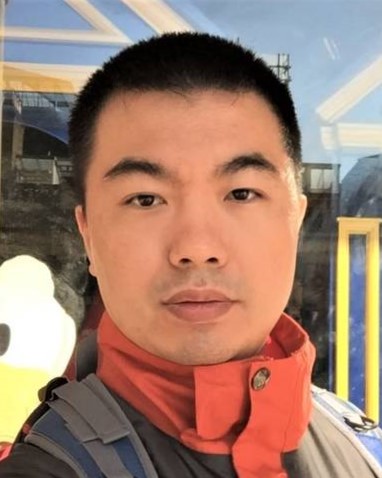}}]{Ming Jiang} is a postdoctoral researcher at the Department of Computer Science and Engineering, University of Minnesota. He obtained his Ph.D. degree in Electrical and Computer Engineering from the National University of Singapore. His M.E and B.E degrees were received from Zhejiang University, Hangzhou, China. His research aim to understand the neural mechanism of attention and to build artificial intelligence systems to simulate where humans look in various contexts. His broader research interests include computer vision, deep neural networks, psychophysics, and cognitive neuroscience.
\end{IEEEbiography}


\begin{IEEEbiography}[{\includegraphics[width=1in,height=1.25in,clip,keepaspectratio]{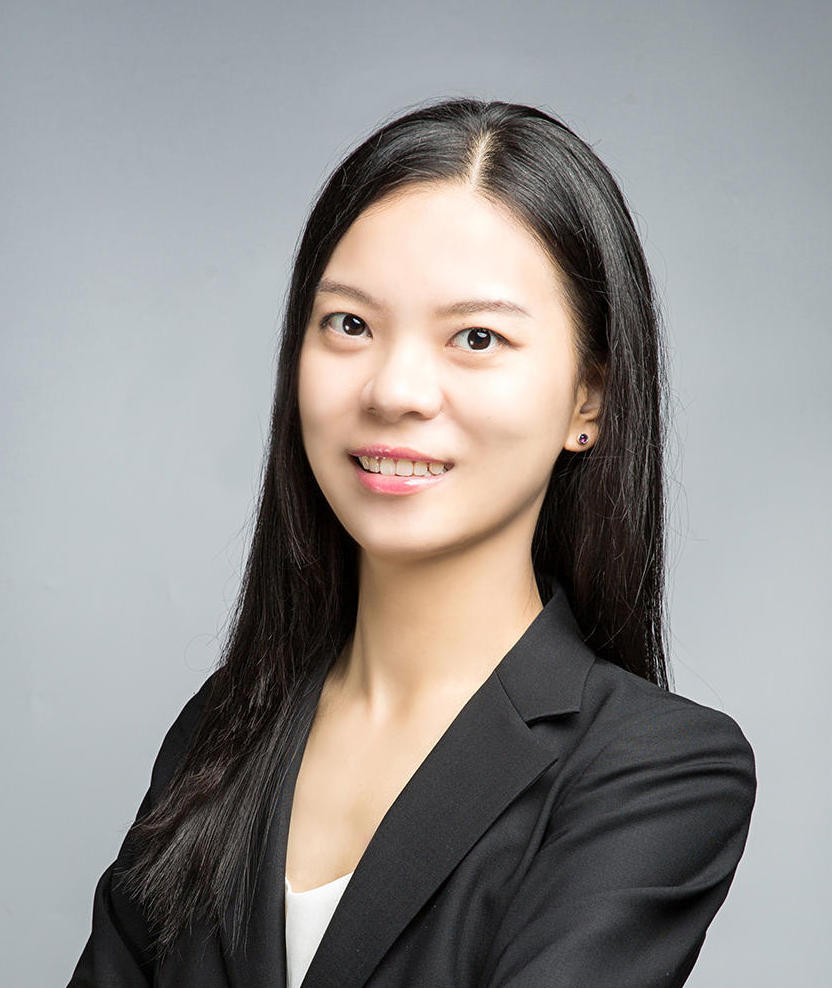}}]{Qi Zhao}
is an assistant professor in the Department of Computer Science and Engineering at the University of Minnesota, Twin Cities.
Her main research interests include computer vision, machine learning, cognitive neuroscience,
and mental disorders. She received her Ph.D.
in computer engineering from the University of
California, Santa Cruz in 2009. She was a postdoctoral researcher in the Computation \& Neural
Systems, and Division of Biology at the California Institute of Technology from 2009 to 2011.
Prior to joining the University of Minnesota, Qi was an assistant professor in the Department of Electrical and Computer Engineering and the
Department of Ophthalmology at the National University of Singapore.
She has published more than 50 journal and conference papers in top
computer vision, machine learning, and cognitive neuroscience venues,
and edited a book with Springer, titled Computational and Cognitive
Neuroscience of Vision, that provides a systematic and comprehensive
overview of vision from various perspectives, ranging from neuroscience
to cognition, and from computational principles to engineering developments. She is a member of the IEEE since 2004.
\end{IEEEbiography}


\vfill


\end{document}